\documentclass[10pt,twocolumn,letterpaper]{article}

\usepackage[pagenumbers,applications]{wacv} % arXiv (de-anonymized, page numbers)
\usepackage{multirow}
\usepackage{wrapfig}
\usepackage{xspace}

\makeatletter
\DeclareRobustCommand\onedot{\futurelet\@let@token\@onedot}
\def\@onedot{\ifx\@let@token.\else.\null\fi\xspace}
\makeatother
\def\eg{\emph{e.g}\onedot}

\def\ie{\emph{i.e}\onedot}

\def\vs{\emph{vs}\onedot}
\def\etal{\emph{et~al}\onedot}

\usepackage{float}

\definecolor{wacvblue}{rgb}{0.21,0.49,0.74}
\usepackage[pagebackref,breaklinks,colorlinks,allcolors=wacvblue]{hyperref}

\usepackage[capitalize,nameinlink,noabbrev]{cleveref}
\crefname{section}{Sec.}{Secs.}
\Crefname{section}{Section}{Sections}
\crefname{figure}{Fig.}{Figs.}
\Crefname{figure}{Figure}{Figures}
\crefname{table}{Tab.}{Tabs.}
\Crefname{table}{Table}{Tables}
\crefname{equation}{Eq.}{Eqs.}
\Crefname{equation}{Equation}{Equations}

\def\wacvPaperID{2245}
\def\confName{WACV}
\def\confYear{2027}

\title{TensorLDM: A Component-Wise Latent Diffusion Model for Volumetric DTI Reconstruction from Sparse DWIs}

\author{
Junhyeok Lee$^{1}$\qquad Kyu Sung Choi$^{2,3,4}$\\[2pt]
$^{1}$Interdisciplinary Program in Cancer Biology, Seoul National University College of Medicine\\
$^{2}$Department of Radiology, Seoul National University Hospital\\
$^{3}$Department of Radiology, Seoul National University College of Medicine\\
$^{4}$Healthcare AI Research Institute, Seoul National University Hospital
}

\begin{document}
\maketitle
\begin{abstract}
Reconstructing diffusion tensors from sparse DWIs is critical for accelerating Diffusion Tensor Imaging (DTI) in clinical settings, yet current deep learning approaches frequently yield anatomically inconsistent or physically implausible tensors.
We introduce TensorLDM, a component-wise latent diffusion model that processes the six tensor components through two group-specific encoders (for diagonal and off-diagonal elements) while maintaining anatomical consistency via shared DWI conditioning.
TensorLDM uses an Anatomy-Conditioned Autoencoder that encourages the latent to focus on tensor properties rather than re-encoding structural information. A shared Cross-Component Attention (CCA) mechanism, applied in both autoencoder refinement and diffusion fine-tuning, models inter-component dependencies, while a Mixture-of-Experts (MoE) DWI conditioner provides component-adaptive conditioning.
On the Human Connectome Project (HCP) dataset under a single-shell, four-volume sparse acquisition, TensorLDM produces the most accurate downstream tractography and tensors with near-ground-truth physical validity (SPD-violation rate 1.54\% \vs\ 1.40\%), with the best or comparable voxel-wise reconstruction accuracy. Geodesic tensor error measured by the Log-Euclidean Metric (LEM) corroborates these gains.
\end{abstract}

\section{Introduction}
\label{sec:intro}

Diffusion Tensor Imaging (DTI) is a Magnetic Resonance Imaging (MRI) technique that quantifies anisotropic water diffusion, enabling non-invasive characterization of white matter microstructure~\cite{basser1994mr, le2001diffusion}. DTI supports neural pathway mapping and extraction of clinically relevant biomarkers across neurological disorders~\cite{behrens2007probabilistic, andica2020mr}. However, high-quality tensor estimation typically requires more than 30 Diffusion-Weighted Images (DWIs) to adequately sample the diffusion signal~\cite{mukherjee2008diffusion}, resulting in prolonged scans that increase patient discomfort and motion sensitivity~\cite{o2011introduction}. Methods that learn empirical priors for reconstructing reliable tensors from substantially fewer DWIs could therefore improve both scan efficiency and diagnostic accuracy.

Reconstructing the six independent components of the diffusion tensor from a sparse set of DWIs is a severely ill-posed inverse problem~\cite{lenglet2009mathematical}. Deep generative models, which have recently advanced medical image reconstruction~\cite{chung2022score}, offer a data-driven approach to this problem~\cite{tian2020deepdti, li2021superdti, zhang2024diff}, yet existing approaches have notable limitations. First, many operate on a 2D slice-by-slice basis, disregarding the inherently three-dimensional continuity of white matter tracts and introducing inter-slice inconsistencies that can propagate into downstream analyses such as tractography~\cite{behrens2007probabilistic}. Second, methods such as SuperDTI~\cite{li2021superdti} and Diff-DTI~\cite{zhang2024diff} synthesize DTI-derived parameter maps such as fractional anisotropy (FA), mean diffusivity (MD), or radial diffusivity (RD) rather than the full tensor, discarding the directional information encoded in the tensor's eigenvectors that is essential for fiber tracking and connectivity analysis.

A further limitation of existing methods is that they treat all six tensor components uniformly through a shared network, ignoring the statistical heterogeneity among them. The diagonal elements ($D_{xx}, D_{yy}, D_{zz}$) represent diffusivity magnitudes along the coordinate axes and generally exhibit higher signal-to-noise ratios and smoother spatial variations, whereas the off-diagonal elements ($D_{xy}, D_{xz}, D_{yz}$) encode cross-directional correlations, are more noise-sensitive, and display more complex spatial patterns. This asymmetry motivates a component-wise architecture.

\begin{figure*}[t]
 \centering
 \includegraphics[width=\linewidth]{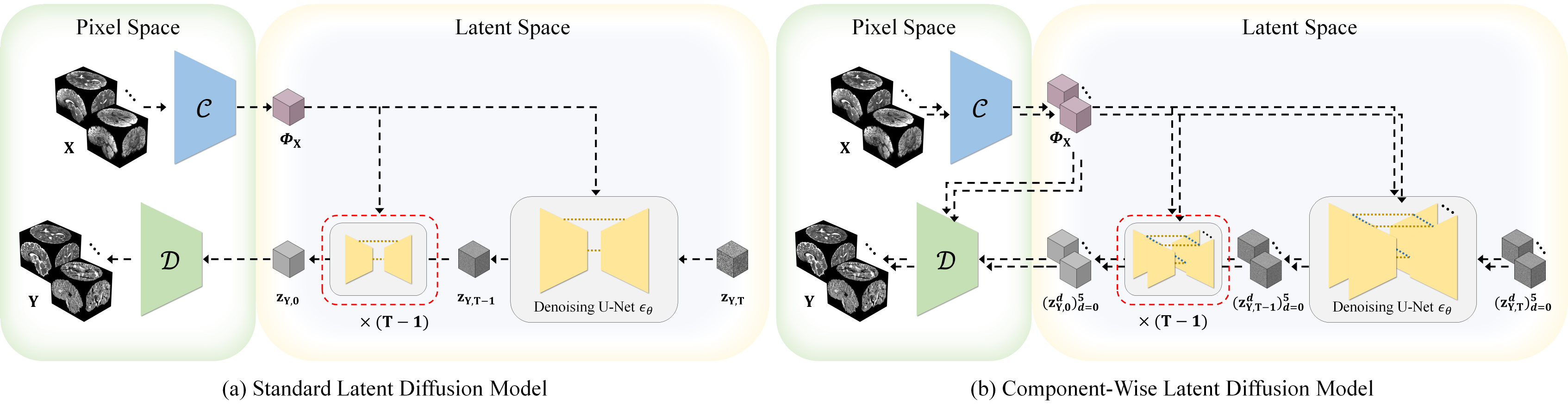}
 \caption{\textbf{Overview of the TensorLDM framework.} (a) A standard latent diffusion model applies the diffusion process to DTI latents $\{\mathbf{z}^d_{\mathbf{Y}}\}$, conditioned on DWI features via concatenation. (b) TensorLDM employs a component-wise latent diffusion model where each tensor component is processed independently with shared anatomical conditioning $\Phi_{\mathbf{X}}$, enabling adaptive component-specific modeling.}
 \label{fig:overview}
\end{figure*}

Beyond these architectural considerations, latent-based generative models face an additional challenge. Multi-stage training has proven effective for stabilizing high-compression latent models~\cite{wang20253d, chendeep}. Building on these insights, we propose TensorLDM, a component-wise latent diffusion model~\cite{ho2020denoising, song2020denoising, rombach2022high} that extends multi-stage optimization to promote tensor-specific properties such as inter-component consistency and symmetric positive definite (SPD) coherence within a volumetric generative framework.

In addition to these modeling challenges, the evaluation of DTI reconstruction itself warrants re-examination. Existing evaluation protocols remain limited in scope. Standard voxel-wise metrics such as Peak Signal-to-Noise Ratio (PSNR) and Structural Similarity Index Measure (SSIM) measure per-element accuracy but do not capture the geometric structure of the tensor field: a high PSNR for individual components does not guarantee that the reconstructed tensors are symmetric positive definite or that their eigenvectors form coherent orientation fields. To provide a more thorough assessment, we complement standard metrics with the manifold-aware Log-Euclidean Metric (LEM)~\cite{arsigny2006log} and downstream probabilistic tractography, evaluating reconstruction quality at the tensor, geometric, and functional levels.

Our key contributions are summarized as follows:
\begin{itemize}
 \item \textbf{Component-Wise Latent Diffusion Model:} We introduce an architecture that employs component-specific encoding and shared anatomical conditioning, enabling specialized processing of diagonal and off-diagonal tensor elements while preserving structural coherence.
 \item \textbf{Anatomy-Conditioned Autoencoder:} We inject DWI features directly into the decoder, encouraging the latent to focus on tensor-specific characteristics rather than re-encoding anatomy, which improves reconstruction fidelity.
 \item \textbf{Cross-Component Attention and Component-Adaptive Conditioning:} We design a CCA mechanism, shared across autoencoder refinement and diffusion fine-tuning, to model inter-component dependencies, and a lightweight Mixture-of-Experts (MoE) conditioner for component-adaptive DWI feature injection.
 \item \textbf{Multi-Level Evaluation:} We complement standard voxel-wise metrics with the manifold-aware LEM and downstream probabilistic tractography, assessing tensor-level geometric consistency beyond conventional image quality measures.
\end{itemize}

\section{Related Work and Background}
\label{sec:related}

\subsection{Diffusion Tensor Model}
DTI models the diffusion process in each voxel using a $3{\times}3$ SPD tensor $\mathbf{D}$~\cite{basser1994mr}. This tensor relates the measured diffusion-weighted signal to the applied diffusion-sensitizing gradients via the Stejskal-Tanner equation~\cite{stejskal1965spin}:
\begin{equation*}
S(\mathbf{g})=S_0 \exp(-b\,\mathbf{g}^\top \mathbf{D}\,\mathbf{g}),
\end{equation*}
where $S_0$ is the non-diffusion-weighted signal intensity, $b$ is the diffusion weighting factor, $\mathbf{g}$ is a unit diffusion gradient direction, and $\mathbf{D}$ is the diffusion tensor. Further details on tensor estimation and derived metrics are provided in~\cref{app:dti}.

Higher-order diffusion models (fiber orientation distributions~\cite{tournier2007robust}, spherical deconvolution, and sparse orientation modeling~\cite{canales2019sparse}) target multi-shell or high-angular-resolution regimes to resolve complex fiber configurations~\cite{karimi2024diffusion}. In contrast, this work focuses on full-tensor reconstruction under an extremely sparse single-shell acquisition, a regime that such higher-order models are generally not designed for.

\subsection{DTI Reconstruction from Sparse Acquisitions}

Reconstructing a diffusion tensor from an insufficient number of DWIs is ill-posed~\cite{tuch2004q}. Early approaches relied on linear or weighted linear least-squares fitting, methods that are computationally simple but highly unstable and sensitive to noise~\cite{basser1994mr}. Model-based approaches used compressed sensing to exploit sparsity priors, with Knoll~\etal~\cite{knoll2015model} applying Total Variation constraints to preserve spatial coherence. More broadly, deep learning has been used to accelerate diffusion MRI by predicting microstructural parameters directly from sparse q-space samples~\cite{golkov2016q} and by image quality transfer for angular and spatial super-resolution~\cite{alexander2017image, tanno2017bayesian}.

SuperDTI~\cite{li2021superdti} demonstrates that convolutional neural networks can map sparse DWIs to diffusion parameter maps from as few as six gradient directions, and FlexDTI~\cite{wu2024flexdti} further incorporates gradient-direction flexibility. Other learning-based methods regress the full tensor or its parameters directly~\cite{tian2020deepdti, li2021superdti}. However, SuperDTI and Diff-DTI~\cite{zhang2024diff} process slices independently, ignoring volumetric anatomical context. Beyond DTI-specific designs, general-purpose conditional image-synthesis models such as adversarial image-to-image translation~\cite{isola2017image, zhu2017unpaired} and transformer-based multimodal medical image synthesis~\cite{dalmaz2022resvit} can be adapted to map DWIs to tensor or parameter maps and serve as strong baselines, yet they are not tailored to the tensor's component structure or its geometric (SPD) constraints.

\subsection{Generative and Latent Diffusion Models for Medical Imaging}

Denoising diffusion probabilistic models~\cite{sohl2015deep, ho2020denoising} now achieve state-of-the-art results in generative modeling, with notable applications in medical imaging, including accelerated MRI reconstruction~\cite{chung2022score} and high-resolution 3D volume synthesis~\cite{wang20253d}.

To reduce the cost of pixel-space diffusion, latent diffusion models (LDMs) carry out the diffusion process in a compact autoencoder latent space~\cite{rombach2022high}. Obtaining a latent space that is simultaneously highly compressed and faithful is difficult, which motivates multi-stage training that decouples representation learning from generative adaptation, as adopted by deep compression autoencoders~\cite{chendeep} and 3D medical latent diffusion models~\cite{wang20253d}. Our framework builds on this multi-stage principle but tailors it to a component-wise tensor latent space, adding tensor-specific objectives (inter-component consistency and symmetric positive definite coherence) that natural-image LDMs do not address.

Within diffusion MRI, several self-supervised methods use diffusion models to restore signal quality from noisy acquisitions~\cite{xiangddm, wuself}. More directly related to our task, Diff-DTI~\cite{zhang2024diff} first applied a diffusion model to rapid DTI reconstruction by conditioning on sparse DWI features to synthesize scalar maps such as FA and MD. In contrast, our approach operates on coupled DWI and tensor latents to directly synthesize the full six-component tensor, from which parameter maps are then derived.

\section{Method: TensorLDM}
\label{sec:method}

\subsection{Problem Definition}

The goal is to reconstruct a full diffusion tensor field from a minimal set of DWIs. The input $\mathbf{X} = \{\mathbf{X}_c\}_{c=0}^3$ consists of four DWI volumes: one non-diffusion-weighted image ($b{=}0$) and three directional DWI volumes, where each $\mathbf{X}_c \in \mathbb{R}^{H \times W \times S}$. The output $\mathbf{Y} = \{\mathbf{Y}_d\}_{d=0}^5$ comprises the six unique elements of the symmetric diffusion tensor ($D_{xx}, D_{yy}, D_{zz}, D_{xy}, D_{xz}, D_{yz}$), where each $\mathbf{Y}_d \in \mathbb{R}^{H \times W \times S}$. Rather than modeling the physical diffusion process, TensorLDM learns an empirical latent prior that captures component-specific relationships between a sparse DWI subset and each tensor component.

\begin{figure}[t]
 \centering
 \includegraphics[width=\linewidth]{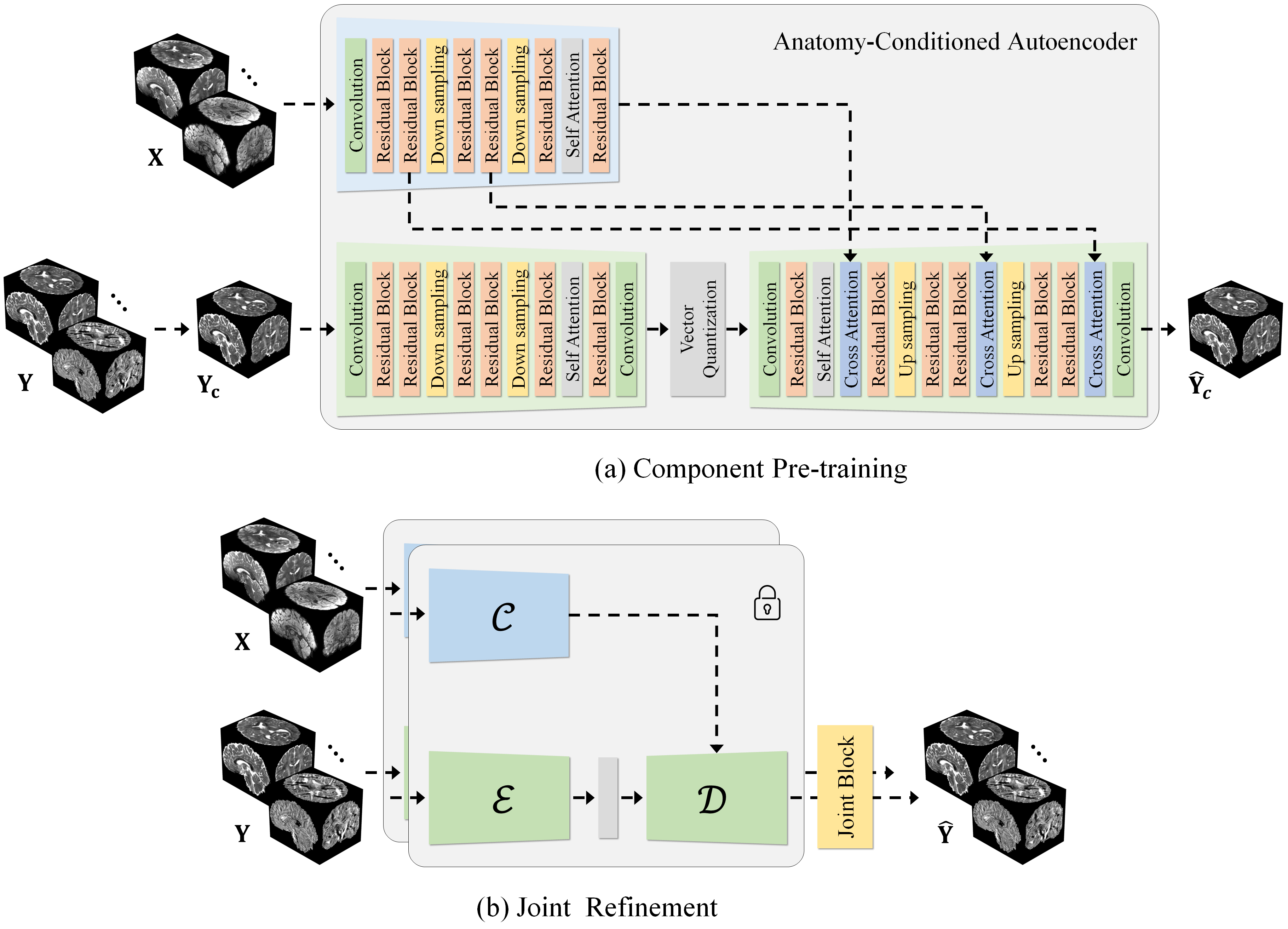}
 \caption{\textbf{Anatomy-Conditioned Autoencoder architecture and training.} (a)~Pre-training and full-volume fine-tuning of the autoencoder. (b)~Joint refinement with CCA and a fusion MLP to promote cross-component coherence.}
 \label{fig:autoencoder}
\end{figure}

\subsection{Overview}

We propose TensorLDM, a variant of the Latent Diffusion Model (LDM)~\cite{rombach2022high} with component-specific pathways and shared DWI conditioning.

\Cref{fig:overview} contrasts our component-wise design with a standard latent diffusion model. Training proceeds in two phases. Phase~I (Stages~1--3) establishes a high-fidelity latent space with an Anatomy-Conditioned Autoencoder (\cref{fig:autoencoder}). Phase~II (Stages~4--5) learns component-wise latent priors conditioned on DWI features (\cref{fig:mtld}). A shared Cross-Component Attention (CCA) mechanism is used in both phases (Stages~3 and~5) to promote inter-component consistency, while Feature-wise Linear Modulation (FiLM)~\cite{perez2018film} provides DWI conditioning throughout.

\subsection{Anatomy-Conditioned Autoencoder for High-Fidelity Latent Representation}

A conventional autoencoder must jointly encode both anatomical structure and tensor properties into a single latent code, creating a bottleneck that limits reconstruction fidelity~\cite{higgins2017beta, chendeep}. Our Anatomy-Conditioned Autoencoder (\cref{fig:autoencoder}) instead supplies anatomical context to the decoder through DWI conditioning, encouraging the latent to allocate its capacity to tensor-specific characteristics rather than re-encoding anatomy. As one indication, a linear probe finds FA far less linearly decodable from the tensor latents ($R^2{=}0.08$) than from the conditioning ($R^2{=}0.61$), consistent with this design intent; we do not claim formal disentanglement (\cref{app:probe}).

Training proceeds in three stages (\cref{fig:autoencoder}a--b): patch-based pre-training, full-volume fine-tuning, and CCA-based joint refinement.

\subsubsection{Autoencoder Architecture and Shared Conditioning (Stages 1 \& 2)}

The autoencoder employs two group-specific encoders (one shared among the three diagonal components and one among the three off-diagonal components), allowing distinct feature representations for each group. A shared conditioner $\mathcal{C}$ extracts multi-scale feature maps $\{\phi_c^{(l)}\}_l$ from each DWI volume $\mathbf{X}_c$. In Phase~I, these per-volume features are uniformly fused across scales to produce a component-independent conditioning representation $\Phi_{\mathbf{X}}$. In Phase~II, they instead serve as input to the MoE DWI conditioner, which produces component-adaptive features $\Phi^d_{\mathbf{X}}$.

During decoding, each component-specific decoder $\mathcal{D}_d$ reconstructs $\hat{\mathbf{Y}}_d$ from latent $\mathbf{z}^d_{\mathbf{Y}}$ and the conditioning features $\Phi_{\mathbf{X}}$. Conditioning features are injected into each component decoder via FiLM layers that apply a learned per-resolution affine transform; in Phase~II $\Phi_{\mathbf{X}}$ is replaced by the component-adaptive $\Phi^d_{\mathbf{X}}$. Swin Transformer attention blocks~\cite{liu2021swin} enable adaptive feature extraction within each decoder.

\subsubsection{Cross-Component Attention and Refinement (Stage 3)}
\label{sec:cca}
The six components must form a valid SPD tensor at each voxel. In the joint refinement step (Stage~3; \cref{fig:autoencoder}b), we upgrade the decoder attention blocks to CCA and fine-tune only these attention blocks, the decoder output layers, and a lightweight joint fusion MLP.

\textbf{Cross-Component Attention (CCA).} A vanilla conditional LDM denoises each tensor component independently and injects DWI features without modeling relationships between components. CCA instead applies multi-head self-attention over a sequence that concatenates all six component latents $\mathbf{Z} = [\mathbf{z}^0_{\mathbf{Y}};\, \ldots;\, \mathbf{z}^5_{\mathbf{Y}}]$ along the component dimension, so that components attend to one another (\eg, $D_{xy}$ attending to $D_{xx}$ and $D_{yy}$), implemented with Swin Transformer blocks~\cite{liu2021swin}. Beyond standard self-attention, CCA introduces two tensor-specific elements. First, learnable per-head direction-mixing matrices $\mathbf{M}_Q, \mathbf{M}_K, \mathbf{M}_V \in \mathbb{R}^{6\times6}$ recombine the projected query, key, and value features across the six tensor components:
\begin{equation*}
\begin{aligned}
\mathbf{Q} &= \mathbf{M}_Q \mathbf{W}_Q \mathbf{Z}, \quad \mathbf{K} = \mathbf{M}_K \mathbf{W}_K \mathbf{Z}, \\
\mathbf{V} &= \mathbf{M}_V \mathbf{W}_V \mathbf{Z},
\end{aligned}
\end{equation*}
where $\mathbf{W}_{\{Q,K,V\}}$ project along the feature dimension and $\mathbf{M}_{\{Q,K,V\}}$ mix along the component dimension. Second, a learnable direction-embedding bias $\mathbf{E}_{\mathrm{dir}}$, whose entry $[\mathbf{E}_{\mathrm{dir}}]_{d,d'}$ scores the geometric relationship between components $d$ and $d'$ (shared across spatial positions), is added to the attention logits alongside the spatial relative-position bias $\mathbf{E}_{\mathrm{pos}}$:
\begin{equation*}
\mathrm{Sim}(\mathbf{Q}, \mathbf{K}) = \frac{\mathbf{Q}\mathbf{K}^\top}{\sqrt{d_k}} + \mathbf{E}_{\mathrm{pos}} + \mathbf{E}_{\mathrm{dir}},
\end{equation*}
where $d_k$ is the key dimension. The same CCA mechanism is reused in Phase~II (Stage~5) to promote cross-component consistency during denoising.

\begin{figure}[t]
 \centering
 \includegraphics[width=\linewidth]{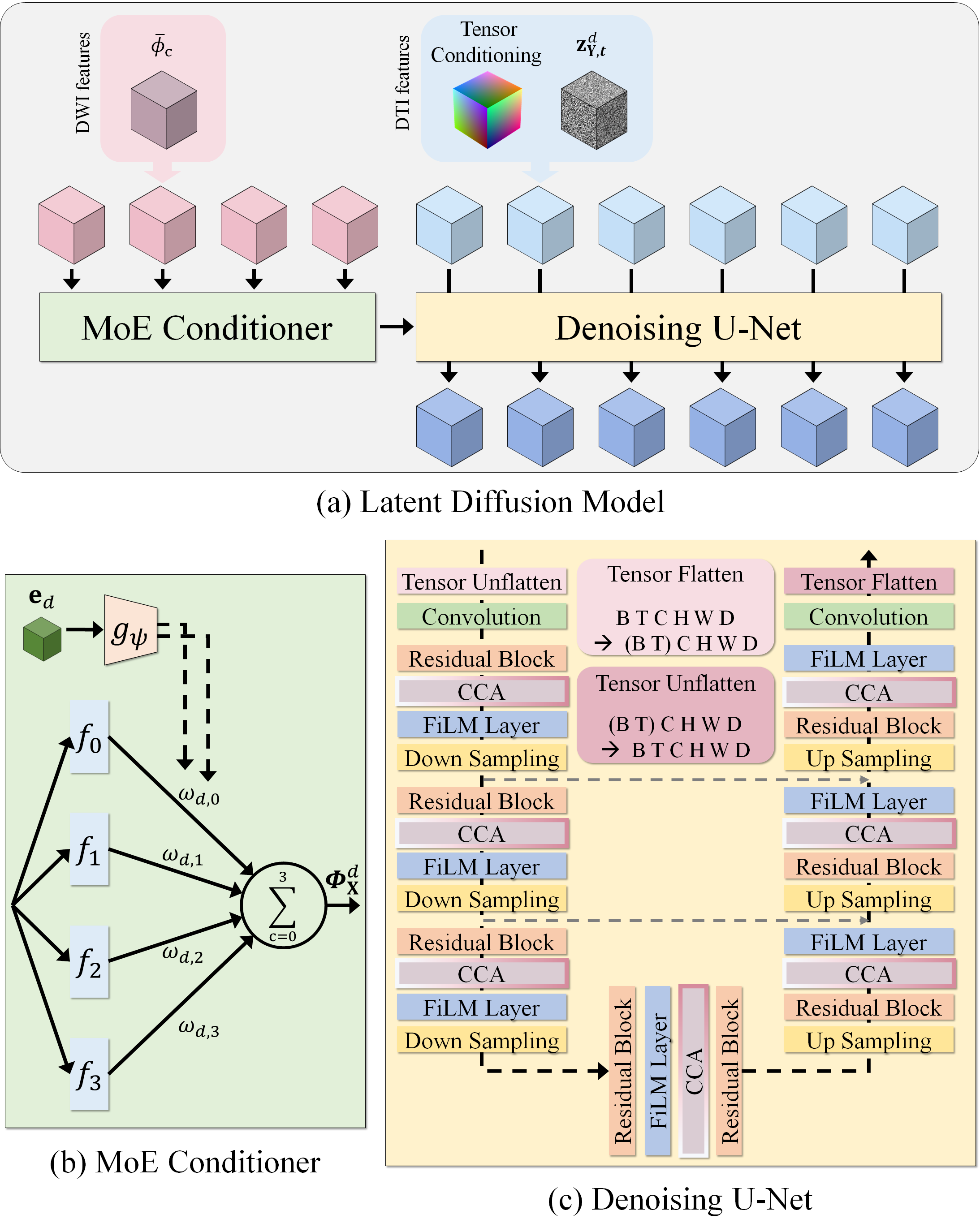}
 \caption{\textbf{Component-wise latent diffusion with MoE DWI conditioner.}
 (a)~MoE conditioner produces component-adaptive $\Phi^d_{\mathbf{X}}$ injected via FiLM.
 (b)~Expert encoders $\{f_c\}$ with gating $g_\psi(\mathbf{e}_d)$ form $\Phi^d_{\mathbf{X}} = \sum_c w_{d,c} \cdot f_c(\bar{\phi}_c)$.
 (c)~Shared U-Net with CCA blocks enabled in Stage~5 only.}
 \label{fig:mtld}
\end{figure}

\begin{figure*}[t]
 \centering
 \includegraphics[width=\linewidth]{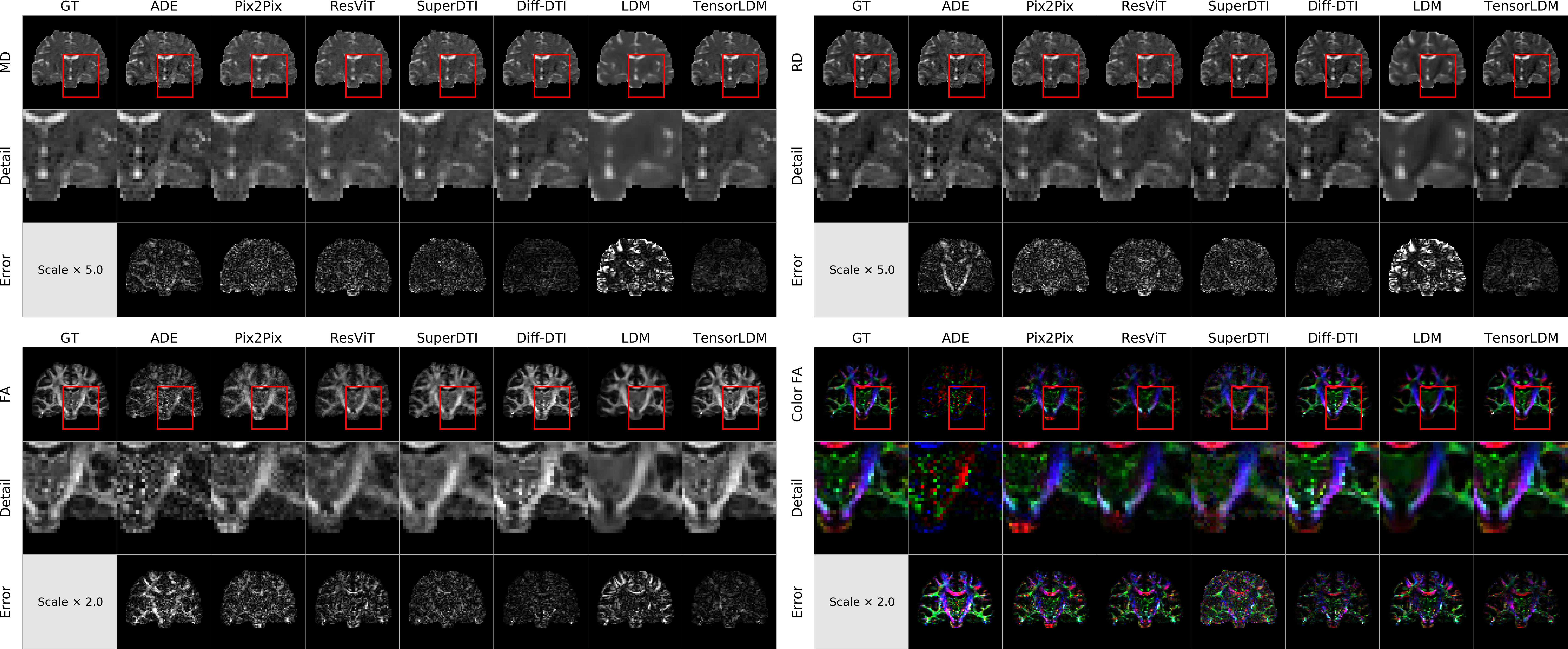}
 \caption{\textbf{Qualitative comparison of DTI parameter maps.} MD, RD, FA, and Color FA (FA modulated by the RGB-encoded principal eigenvector orientation) on a coronal slice. TensorLDM produces reconstructions closest to GT. Magnified insets and error maps highlight improved detail recovery.}
 \label{fig:qual_params_coronal}
\end{figure*}

\textbf{Training Objectives.} Stages~1 and~2 minimize a voxel-wise $\ell_1$ reconstruction loss $\mathcal{L}_{\mathrm{rec}} = \sum_{d=0}^{5} \lVert \hat{\mathbf{Y}}_d - \mathbf{Y}_d \rVert_1$ with a vector-quantization (VQ) commitment loss $\mathcal{L}_{\mathrm{VQ}}$~\cite{van2017neural}. Stage~3 augments this with a Log-Euclidean loss to encourage SPD consistency:
\begin{equation*}
\mathcal{L}_{\mathrm{AE}} = \mathcal{L}_{\mathrm{rec}} + \mathcal{L}_{\mathrm{VQ}} + \lambda_{\mathrm{LEM}} \lVert \mathrm{Log}(\hat{\mathbf{D}}) - \mathrm{Log}(\mathbf{D}) \rVert_F,
\end{equation*}
where $\mathrm{Log}(\cdot)$ denotes the matrix logarithm, $\hat{\mathbf{D}}$ and $\mathbf{D}$ are the $3{\times}3$ tensors assembled from $\{\hat{\mathbf{Y}}_d\}$ and $\{\mathbf{Y}_d\}$, and $\lambda_{\mathrm{LEM}}$ balances component-specific fidelity with manifold-aware geometric constraints. The term $\lVert \mathrm{Log}(\hat{\mathbf{D}}) - \mathrm{Log}(\mathbf{D}) \rVert_F$ (averaged over brain voxels) is the Log-Euclidean Metric (LEM)~\cite{arsigny2006log}, the geodesic distance between SPD tensors on the log-Euclidean manifold (\cref{app:metrics}). Full architectural details are provided in~\cref{app:impl}.

\subsection{Component-Wise Latent Diffusion}

While the Anatomy-Conditioned Autoencoder provides high-fidelity reconstruction when ground-truth latents are available, it cannot model the distribution of plausible tensors given only sparse DWI inputs. To bridge this gap, Phase~II learns to generate tensor component latents conditioned on sparse DWI features (\cref{fig:mtld}). Let $\mathbf{z}^d_{\mathbf{Y}}$ denote the latent for tensor component $d \in \{0,\dots,5\}$. The forward diffusion process adds Gaussian noise independently to each component latent over $T$ timesteps:
\begin{equation*}
\mathbf{z}^d_{\mathbf{Y},t} = \sqrt{\bar{\alpha}_t}\mathbf{z}^d_{\mathbf{Y},0} + \sqrt{1-\bar{\alpha}_t}\boldsymbol{\epsilon}^d,
\end{equation*}
where $\{\beta_s\}_{s=1}^{T}$ is a predefined noise schedule, $T$ is the total number of diffusion steps, $\bar{\alpha}_t = \prod_{s=1}^t (1-\beta_s)$, and $\boldsymbol{\epsilon}^d \sim \mathcal{N}(\mathbf{0}, \mathbf{I})$ is sampled independently per component.

\subsubsection{Denoising Backbone and MoE Conditioner}

\paragraph{Shared Denoising Backbone.}
The entire Anatomy-Conditioned Autoencoder (encoders, decoders, and conditioner) is frozen during Phase~II; only the diffusion model and its conditioning modules are trained. A single shared U-Net $\boldsymbol{\epsilon}_\theta$ serves as the denoising backbone for all tensor components (\cref{fig:mtld}a,c). To identify which component is being denoised, a learnable embedding $\mathbf{e}_d = \mathrm{Embed}(d)$ is concatenated with the noisy latent to form the augmented input $\tilde{\mathbf{z}}^d_{\mathbf{Y},t} = [\mathbf{z}^d_{\mathbf{Y},t};\, \mathbf{e}_d]$. The backbone receives component-adaptive conditioning features $\Phi^d_{\mathbf{X}}$ via FiLM layers at each resolution level. These features are produced by a dedicated MoE DWI conditioner described below.

\paragraph{MoE DWI Conditioner.}
\label{sec:moe}
Inspired by mixture-of-experts architectures~\cite{shazeer2017outrageously, riquelme2021scaling}, we introduce a lightweight MoE DWI conditioner $\mathcal{C}_{\mathrm{MoE}}$ that produces component-adaptive conditioning features $\Phi^d_{\mathbf{X}}$, which are injected into the shared denoising backbone via FiLM (\cref{fig:mtld}a). Because the four DWI volumes carry asymmetric directional information, the MoE conditioner learns to selectively weight them for each tensor component rather than averaging uniformly.

Using the multi-scale feature maps $\{\phi_c^{(l)}\}_l$ extracted by the shared conditioner $\mathcal{C}$, the MoE conditioner concatenates and downsamples them to obtain per-volume representations $\bar{\phi}_c$. Each $\bar{\phi}_c$ is then processed by a dedicated DWI expert encoder $f_c$, and a gating network produces component-specific combination weights (\cref{fig:mtld}b):
\begin{equation*}
\mathbf{w}_d = \mathrm{softmax}\bigl(g_\psi(\mathbf{e}_d)\bigr), \quad \Phi^d_{\mathbf{X}} = \textstyle\sum_{c=0}^{3} w_{d,c} \cdot f_c(\bar{\phi}_c),
\end{equation*}
where $g_\psi$ is a lightweight gating network and $\mathbf{e}_d$ is the component embedding shared with the backbone input.

\begin{table*}[t]
 \caption{\textbf{Quantitative comparison of DTI parameter maps.} Normalized Mean Squared Error (NMSE), PSNR, and SSIM (mean$_{\text{\tiny std}}$) over the 195-subject test set. Best in \textbf{bold}, second-best \underline{underlined}.}
 \label{tab:quant_params}
 \centering
 \scriptsize
 \resizebox{\textwidth}{!}{%
 \begin{tabular}{lcccccccccccc}
 \toprule
 \textbf{Model}
 & \multicolumn{3}{c}{\textbf{MD}}
 & \multicolumn{3}{c}{\textbf{RD}}
 & \multicolumn{3}{c}{\textbf{FA}}
 & \multicolumn{3}{c}{\textbf{Color FA}} \\
 \cmidrule(lr){2-4}
 \cmidrule(lr){5-7}
 \cmidrule(lr){8-10}
 \cmidrule(lr){11-13}
 & NMSE & PSNR & SSIM
 & NMSE & PSNR & SSIM
 & NMSE & PSNR & SSIM
 & NMSE & PSNR & SSIM \\
 \midrule

 ADE
 & \underline{0.02}$_{\text{\tiny 0.00}}$ & 28.97$_{\text{\tiny 0.66}}$ & \underline{0.97}$_{\text{\tiny 0.01}}$
 & \underline{0.02}$_{\text{\tiny 0.00}}$ & 28.74$_{\text{\tiny 0.85}}$ & 0.96$_{\text{\tiny 0.01}}$
 & 0.37$_{\text{\tiny 0.03}}$ & 16.26$_{\text{\tiny 0.47}}$ & 0.75$_{\text{\tiny 0.02}}$
 & 1.07$_{\text{\tiny 0.03}}$ & 19.94$_{\text{\tiny 0.52}}$ & 0.60$_{\text{\tiny 0.03}}$ \\

 CycleGAN
 & 0.13$_{\text{\tiny 0.01}}$ & 19.85$_{\text{\tiny 0.54}}$ & 0.80$_{\text{\tiny 0.02}}$
 & 0.16$_{\text{\tiny 0.01}}$ & 20.06$_{\text{\tiny 0.80}}$ & 0.80$_{\text{\tiny 0.02}}$
 & 0.50$_{\text{\tiny 0.04}}$ & 14.93$_{\text{\tiny 0.48}}$ & 0.62$_{\text{\tiny 0.03}}$
 & 1.08$_{\text{\tiny 0.10}}$ & 19.93$_{\text{\tiny 0.75}}$ & 0.60$_{\text{\tiny 0.03}}$ \\

 Pix2Pix
 & 0.04$_{\text{\tiny 0.00}}$ & 24.98$_{\text{\tiny 0.56}}$ & 0.93$_{\text{\tiny 0.01}}$
 & 0.05$_{\text{\tiny 0.01}}$ & 25.08$_{\text{\tiny 0.82}}$ & 0.93$_{\text{\tiny 0.01}}$
 & 0.29$_{\text{\tiny 0.02}}$ & 17.25$_{\text{\tiny 0.49}}$ & 0.76$_{\text{\tiny 0.02}}$
 & 0.79$_{\text{\tiny 0.06}}$ & 21.28$_{\text{\tiny 0.69}}$ & 0.71$_{\text{\tiny 0.02}}$ \\

 ResViT
 & 0.03$_{\text{\tiny 0.00}}$ & 25.69$_{\text{\tiny 0.57}}$ & 0.94$_{\text{\tiny 0.01}}$
 & 0.04$_{\text{\tiny 0.00}}$ & 25.76$_{\text{\tiny 0.80}}$ & 0.94$_{\text{\tiny 0.01}}$
 & 0.28$_{\text{\tiny 0.02}}$ & 17.51$_{\text{\tiny 0.44}}$ & 0.76$_{\text{\tiny 0.02}}$
 & \underline{0.75}$_{\text{\tiny 0.04}}$ & 21.49$_{\text{\tiny 0.64}}$ & 0.72$_{\text{\tiny 0.02}}$ \\

 SuperDTI
 & 0.04$_{\text{\tiny 0.01}}$ & 25.53$_{\text{\tiny 0.67}}$ & 0.94$_{\text{\tiny 0.01}}$
 & 0.05$_{\text{\tiny 0.01}}$ & 25.38$_{\text{\tiny 0.88}}$ & 0.94$_{\text{\tiny 0.01}}$
 & \underline{0.22}$_{\text{\tiny 0.02}}$ & 18.54$_{\text{\tiny 0.47}}$ & 0.81$_{\text{\tiny 0.02}}$
 & \underline{0.75}$_{\text{\tiny 0.04}}$ & 21.50$_{\text{\tiny 0.55}}$ & 0.69$_{\text{\tiny 0.02}}$ \\

 Diff-DTI
 & \textbf{0.01}$_{\text{\tiny 0.00}}$ & \underline{32.66}$_{\text{\tiny 0.63}}$ & \underline{0.97}$_{\text{\tiny 0.00}}$
 & \textbf{0.01}$_{\text{\tiny 0.00}}$ & \underline{32.76}$_{\text{\tiny 0.80}}$ & \underline{0.97}$_{\text{\tiny 0.00}}$
 & \textbf{0.08}$_{\text{\tiny 0.01}}$ & \underline{22.71}$_{\text{\tiny 0.38}}$ & \underline{0.93}$_{\text{\tiny 0.00}}$
 & \textbf{0.33}$_{\text{\tiny 0.02}}$ & \underline{24.60}$_{\text{\tiny 0.56}}$ & \underline{0.89}$_{\text{\tiny 0.01}}$ \\

 LDM
 & 0.13$_{\text{\tiny 0.01}}$ & 20.02$_{\text{\tiny 0.63}}$ & 0.81$_{\text{\tiny 0.02}}$
 & 0.15$_{\text{\tiny 0.01}}$ & 20.19$_{\text{\tiny 0.84}}$ & 0.82$_{\text{\tiny 0.02}}$
 & 0.37$_{\text{\tiny 0.02}}$ & 16.28$_{\text{\tiny 0.45}}$ & 0.67$_{\text{\tiny 0.03}}$
 & 0.78$_{\text{\tiny 0.03}}$ & 21.36$_{\text{\tiny 0.64}}$ & 0.68$_{\text{\tiny 0.03}}$ \\

 TensorLDM
 & \textbf{0.01}$_{\text{\tiny 0.00}}$ & \textbf{32.86}$_{\text{\tiny 0.82}}$ & \textbf{0.99}$_{\text{\tiny 0.00}}$
 & \textbf{0.01}$_{\text{\tiny 0.00}}$ & \textbf{33.15}$_{\text{\tiny 0.96}}$ & \textbf{0.99}$_{\text{\tiny 0.00}}$
 & \textbf{0.08}$_{\text{\tiny 0.01}}$ & \textbf{23.07}$_{\text{\tiny 0.44}}$ & \textbf{0.94}$_{\text{\tiny 0.00}}$
 & \textbf{0.33}$_{\text{\tiny 0.02}}$ & \textbf{25.03}$_{\text{\tiny 0.60}}$ & \textbf{0.90}$_{\text{\tiny 0.01}}$ \\

 \bottomrule
 \end{tabular}
 }
\end{table*}

\begin{table*}[t]
 \caption{\textbf{Quantitative comparison of diffusion tensor components.} PSNR, SSIM, and LEM (mean$_{\text{\tiny std}}$) across six tensor components, over the 195-subject test set. Best in \textbf{bold}, second-best \underline{underlined}.}
 \label{tab:quant_tensors}
 \centering
 \scriptsize
 \resizebox{\textwidth}{!}{%
 \begin{tabular}{l*{12}{c}c}\toprule
 \textbf{Model} & \multicolumn{2}{c}{\textbf{$D_{xx}$}} & \multicolumn{2}{c}{\textbf{$D_{yy}$}} & \multicolumn{2}{c}{\textbf{$D_{zz}$}}
 & \multicolumn{2}{c}{\textbf{$D_{xy}$}} & \multicolumn{2}{c}{\textbf{$D_{xz}$}} & \multicolumn{2}{c}{\textbf{$D_{yz}$}} & \multirow{2}{*}{\textbf{LEM}} \\
 \cmidrule(lr){2-3}\cmidrule(lr){4-5}\cmidrule(lr){6-7}\cmidrule(lr){8-9}\cmidrule(lr){10-11}\cmidrule(lr){12-13}
 & PSNR & SSIM & PSNR & SSIM & PSNR & SSIM & PSNR & SSIM & PSNR & SSIM & PSNR & SSIM & \\
 \midrule
 ADE & 29.71$_{\text{\tiny 0.41}}$ & 0.92$_{\text{\tiny 0.01}}$ & 30.22$_{\text{\tiny 0.74}}$ & 0.93$_{\text{\tiny 0.01}}$ & 29.84$_{\text{\tiny 0.56}}$ & 0.92$_{\text{\tiny 0.01}}$ & 24.03$_{\text{\tiny 1.21}}$ & 0.64$_{\text{\tiny 0.04}}$ & 23.39$_{\text{\tiny 1.52}}$ & 0.64$_{\text{\tiny 0.04}}$ & 23.43$_{\text{\tiny 1.50}}$ & 0.63$_{\text{\tiny 0.04}}$ & 0.65$_{\text{\tiny 0.03}}$ \\

 CycleGAN & 25.25$_{\text{\tiny 0.36}}$ & 0.81$_{\text{\tiny 0.02}}$ & 25.20$_{\text{\tiny 0.46}}$ & 0.82$_{\text{\tiny 0.02}}$ & 25.22$_{\text{\tiny 0.48}}$ & 0.81$_{\text{\tiny 0.01}}$ & 21.41$_{\text{\tiny 1.23}}$ & 0.60$_{\text{\tiny 0.04}}$ & 21.25$_{\text{\tiny 1.56}}$ & 0.60$_{\text{\tiny 0.04}}$ & 22.80$_{\text{\tiny 1.72}}$ & 0.65$_{\text{\tiny 0.05}}$ & 1.00$_{\text{\tiny 0.09}}$ \\

 Pix2Pix & 29.92$_{\text{\tiny 0.38}}$ & 0.93$_{\text{\tiny 0.01}}$ & 30.02$_{\text{\tiny 0.49}}$ & 0.93$_{\text{\tiny 0.01}}$ & 30.05$_{\text{\tiny 0.48}}$ & 0.93$_{\text{\tiny 0.01}}$ & 24.87$_{\text{\tiny 1.27}}$ & 0.70$_{\text{\tiny 0.04}}$ & 24.30$_{\text{\tiny 1.65}}$ & 0.69$_{\text{\tiny 0.04}}$ & 26.19$_{\text{\tiny 1.64}}$ & 0.77$_{\text{\tiny 0.03}}$ & 0.66$_{\text{\tiny 0.05}}$ \\

 ResViT & 30.85$_{\text{\tiny 0.38}}$ & 0.94$_{\text{\tiny 0.01}}$ & 30.93$_{\text{\tiny 0.48}}$ & 0.94$_{\text{\tiny 0.01}}$ & 30.97$_{\text{\tiny 0.49}}$ & 0.94$_{\text{\tiny 0.01}}$ & 25.94$_{\text{\tiny 1.24}}$ & \underline{0.73}$_{\text{\tiny 0.03}}$ & 25.61$_{\text{\tiny 1.55}}$ & 0.73$_{\text{\tiny 0.03}}$ & \underline{27.26}$_{\text{\tiny 1.59}}$ & \underline{0.79}$_{\text{\tiny 0.03}}$ & 0.57$_{\text{\tiny 0.03}}$ \\

 SuperDTI & 30.49$_{\text{\tiny 0.43}}$ & 0.94$_{\text{\tiny 0.01}}$ & 30.46$_{\text{\tiny 0.62}}$ & 0.94$_{\text{\tiny 0.01}}$ & 30.56$_{\text{\tiny 0.52}}$ & 0.94$_{\text{\tiny 0.01}}$ & 20.84$_{\text{\tiny 1.23}}$ & 0.61$_{\text{\tiny 0.04}}$ & 20.36$_{\text{\tiny 1.55}}$ & 0.61$_{\text{\tiny 0.04}}$ & 19.75$_{\text{\tiny 1.54}}$ & 0.58$_{\text{\tiny 0.04}}$ & 0.71$_{\text{\tiny 0.05}}$ \\

 Diff-DTI & \underline{35.86}$_{\text{\tiny 0.46}}$ & \underline{0.98}$_{\text{\tiny 0.00}}$ & \underline{36.27}$_{\text{\tiny 0.60}}$ & \underline{0.98}$_{\text{\tiny 0.00}}$ & \underline{36.16}$_{\text{\tiny 0.52}}$ & \underline{0.98}$_{\text{\tiny 0.00}}$ & 21.22$_{\text{\tiny 1.24}}$ & 0.65$_{\text{\tiny 0.03}}$ & 20.64$_{\text{\tiny 1.55}}$ & 0.65$_{\text{\tiny 0.03}}$ & 19.75$_{\text{\tiny 1.52}}$ & 0.61$_{\text{\tiny 0.04}}$ & \underline{0.49}$_{\text{\tiny 0.02}}$ \\

 LDM & 26.32$_{\text{\tiny 0.50}}$ & 0.85$_{\text{\tiny 0.01}}$ & 26.13$_{\text{\tiny 0.52}}$ & 0.85$_{\text{\tiny 0.01}}$ & 26.24$_{\text{\tiny 0.56}}$ & 0.85$_{\text{\tiny 0.01}}$ & \underline{26.18}$_{\text{\tiny 1.23}}$ & \textbf{0.76}$_{\text{\tiny 0.03}}$ & \underline{25.66}$_{\text{\tiny 1.52}}$ & \underline{0.75}$_{\text{\tiny 0.03}}$ & 25.53$_{\text{\tiny 1.51}}$ & 0.75$_{\text{\tiny 0.04}}$ & 0.70$_{\text{\tiny 0.03}}$ \\

 TensorLDM & \textbf{36.45}$_{\text{\tiny 0.52}}$ & \textbf{0.99}$_{\text{\tiny 0.00}}$ & \textbf{36.58}$_{\text{\tiny 0.66}}$ & \textbf{0.99}$_{\text{\tiny 0.00}}$ & \textbf{36.75}$_{\text{\tiny 0.59}}$ & \textbf{0.99}$_{\text{\tiny 0.00}}$ & \textbf{29.30}$_{\text{\tiny 1.20}}$ & 0.72$_{\text{\tiny 0.04}}$ & \textbf{29.79}$_{\text{\tiny 1.61}}$ & \textbf{0.82}$_{\text{\tiny 0.04}}$ & \textbf{30.11}$_{\text{\tiny 1.67}}$ & \textbf{0.83}$_{\text{\tiny 0.05}}$ & \textbf{0.33}$_{\text{\tiny 0.02}}$ \\

 \bottomrule
 \end{tabular}
 }
\end{table*}

\subsubsection{Two-Stage Diffusion Training (Stages 4 \& 5)}

\paragraph{Stage 4: Independent Component Pre-training.}
We first pre-train $\boldsymbol{\epsilon}_\theta$ in an independent setting where each component is processed in isolation, with cross-component attention deactivated. The objective is to predict the noise added to each component:
\begin{equation*}
\mathcal{L}_{\mathrm{pretrain}} = \mathbb{E}_{t, d, \mathbf{z}^d_{\mathbf{Y},0}, \boldsymbol{\epsilon}^d} \left[ \lVert\boldsymbol{\epsilon}^d - \boldsymbol{\epsilon}_{\theta}(\tilde{\mathbf{z}}^d_{\mathbf{Y},t}, t, \Phi^d_{\mathbf{X}}, d)\rVert_2^2 \right].
\end{equation*}

\paragraph{Stage 5: Joint Component Fine-tuning.}
Following pre-training, we activate the CCA blocks within the U-Net attention layers (\cref{fig:mtld}c) to enable joint synthesis of all six components. For parameter efficiency, only the attention parameters (projection weights, relative position bias, direction embeddings, mixing matrices) and FiLM conditioning modules are fine-tuned. All other U-Net parameters remain frozen. For component $d$, the noise prediction from the full denoising network is:
\begin{equation*}
\hat{\boldsymbol{\epsilon}}^d = \boldsymbol{\epsilon}_\theta\!\left(\{\tilde{\mathbf{z}}^{d'}_{\mathbf{Y},t}\}_{d'=0}^{5},\, t,\, \Phi^d_{\mathbf{X}},\, d\right).
\end{equation*}
The joint training objective sums over all components:
\begin{equation*}
\mathcal{L}_{\mathrm{cond}} = \mathbb{E}_{t, \{\mathbf{z}^d_{\mathbf{Y},0}\}_d, \{\boldsymbol{\epsilon}^d\}_d} \!\left[ \sum_{d=0}^5 \left\lVert\boldsymbol{\epsilon}^d - \hat{\boldsymbol{\epsilon}}^d\right\rVert_2^2 \right]\!.
\end{equation*}

\section{Experiments}
\label{sec:experiments}

\subsection{Experimental Setup}

\subsubsection{Data and Preprocessing}

All experiments use diffusion MRI data from the HCP Young Adult dataset~\cite{van2013wu}, preprocessed with the standard HCP pipelines~\cite{glasser2013minimal}. We use single-shell DWI volumes at $b{=}1000$~s/mm$^2$, resampled to 2\,mm isotropic resolution. Ground-truth diffusion tensors are computed via linear least-squares fitting from the full acquisition (18 $b{=}0$ volumes and 90 DWI directions). The input is a sparse 4-volume DWI stack, and the output is the complete 6-component tensor field. Three-direction trace/ADC DWI is a standard clinical vendor protocol chosen to keep scan time short~\cite{havsteen2017comparison}, so our four-volume input reflects a realistic clinical operating point rather than an artificial worst case. The dataset of 973 subjects is split into training (681), validation (97), and test (195) sets. Further details are in~\cref{app:data}.

\subsubsection{Implementation Details}

All experiments are implemented in PyTorch~\cite{paszke2019pytorch} and conducted on a single NVIDIA A6000 GPU. Training follows the five-stage pipeline. Detailed architectures, loss functions, and stage-specific hyperparameters are provided in~\cref{app:impl}.

\subsubsection{Competing Methods}

We benchmark TensorLDM against seven baselines: Analytic Diagonal Estimation (ADE), a non-learning baseline assuming a diagonal tensor; CycleGAN~\cite{zhu2017unpaired}; Pix2Pix~\cite{isola2017image}; ResViT~\cite{dalmaz2022resvit}; SuperDTI~\cite{li2021superdti}; Diff-DTI~\cite{zhang2024diff}; and a vanilla conditional LDM~\cite{rombach2022high}. All learning-based methods use 3D networks trained on the same data splits and inputs, except SuperDTI and Diff-DTI, which follow their original 2D slice-based protocols. We evaluate these two methods as published rather than re-engineering them into 3D six-component regressors, so the comparison reflects their scalar-map design rather than our evaluation protocol. Since SuperDTI and Diff-DTI predict parameter maps rather than the full tensor, we reconstruct approximate tensors under a cylindrically symmetric model to enable tensor-domain evaluation (LEM, tractography). Details and a fairness discussion are in~\cref{app:recon_from_params}. Full method descriptions are in~\cref{app:baselines}.

\subsection{Main Results}
\subsubsection{Evaluation Results across Voxel, Geometric, and Functional Levels}

\paragraph{Voxel-Wise Accuracy and Parameter Maps.}
We evaluate standard DTI-derived parameter maps, namely Mean Diffusivity (MD), Radial Diffusivity (RD), Fractional Anisotropy (FA), and Color FA, to assess voxel-wise reconstruction fidelity (\cref{tab:quant_params,fig:qual_params_coronal}). TensorLDM achieves the best or tied-best results across all four parameter maps. Among learning-based baselines, Diff-DTI (which also utilizes a conditional diffusion model) achieves competitive scalar parameter accuracy. However, because Diff-DTI operates on a 2D slice-by-slice basis in its original formulation, it exhibits noticeable inter-slice discontinuities in coronal views (\cref{fig:qual_params_coronal}), whereas TensorLDM's volumetric formulation maintains spatial continuity. The difference in FA NMSE between TensorLDM and Diff-DTI is not statistically significant ($p{=}0.097$), indicating that both models capture general diffusion anisotropy magnitudes comparably.

\paragraph{Tensor Geometry and Physical Validity.}
Reconstructing the full six-component tensor field is more challenging than regressing scalar parameter maps because the individual components must adhere to the symmetric positive definite (SPD) manifold. Voxel-wise metrics on individual tensor components (\cref{tab:quant_tensors}) show that TensorLDM outperforms all baselines on nearly every tensor component (the main exception being $D_{xy}$ SSIM, discussed below), with the largest gains on the noisier off-diagonal elements ($D_{xy}, D_{xz}, D_{yz}$), which capture cross-directional diffusion correlations. Baselines that regress parameter maps (SuperDTI, Diff-DTI) discard true directional heterogeneity, yielding poor off-diagonal accuracy and high LEM errors (\cref{app:recon_from_params}). 

Furthermore, conventional voxel-wise metrics can be misleading: the vanilla LDM baseline achieves a slightly higher SSIM on $D_{xy}$ (0.76 \vs\ 0.72) than TensorLDM, but at the expense of significantly lower diagonal-component accuracy and a much higher LEM (0.70 \vs\ 0.33), demonstrating that high independent voxel-wise similarity does not guarantee tensor-level geometric coherence. Manifold-level accuracy is also reflected in the SPD violation rate, the percentage of voxels containing negative eigenvalues (\cref{tab:spd_violation}). While the ground-truth linear least-squares fit from 90 directions exhibits a 1.40\% violation rate due to noise, TensorLDM yields a physically plausible violation rate of 1.54\% (which degrades to 1.61\% without the soft LEM constraint $\mathcal{L}_{\mathrm{LEM}}$), consistent with the soft geometric constraint contributing to positive-definiteness preservation.

\begin{figure}[t]
 \centering
 \includegraphics[width=\linewidth]{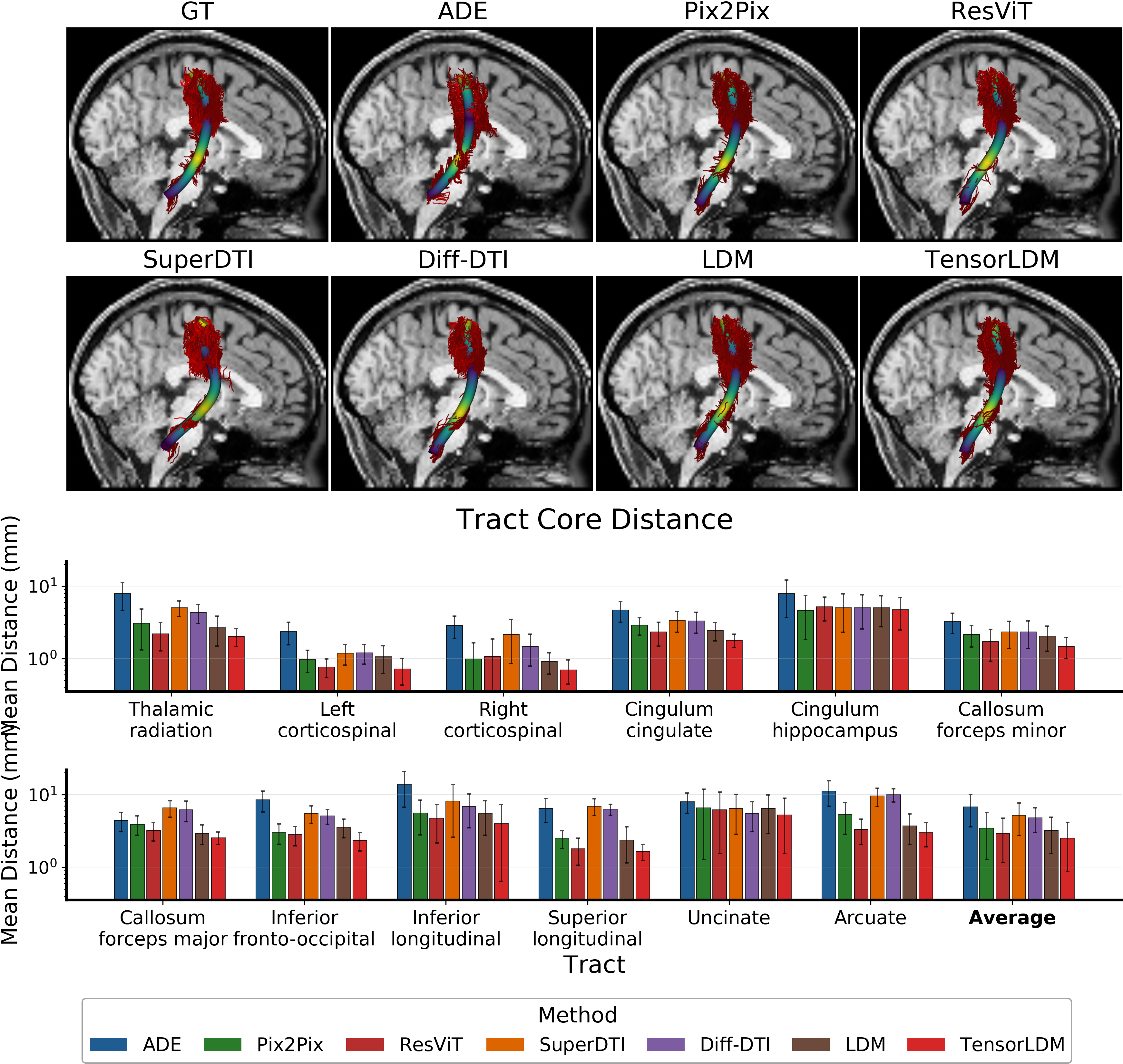}
 \caption{\textbf{Tractography comparison.} (Top) Right corticospinal tract (CST) reconstructions per method. TensorLDM tracts most closely match the ground truth. SuperDTI and Diff-DTI use approximate tensors from cylindrical symmetry reconstruction. (Bottom) Mean tract core distance (mm, log scale) across 12 bundles confirms TensorLDM yields the most accurate fiber tracking.}
 \label{fig:tractography}
\end{figure}

\paragraph{Downstream Tractography and Clinical Utility.}
To evaluate how these geometric improvements translate to clinical utility, we perform whole-brain probabilistic tractography, which is highly sensitive to the principal diffusion orientation (first eigenvector of the tensor). Errors in off-diagonal components propagate directly during streamline generation, leading to anatomically incorrect fiber tracts. As illustrated in \cref{fig:tractography} (top), the reconstructed right corticospinal tract (CST) from TensorLDM matches the 90-direction ground truth most closely. This is quantitatively supported by the mean Tract Core Distance (TCD) measured across 12 major white matter bundles (\cref{fig:tractography}, bottom), where TensorLDM achieves the lowest discrepancy relative to the ground truth, validating the downstream utility of the reconstructed tensor fields.

\begin{table*}[t]
 \caption{\textbf{Ablation study on TensorLDM.}
 Variants are grouped by (a) architectural design and geometric constraint, and (b) multi-stage training.
 Values are mean$_{\text{\tiny std}}$ for PSNR (tensor components and parameter maps) and LEM ($\downarrow$). Bold marks the best score.
 }
 \label{tab:ablation_ae_params}
 \centering
 \scriptsize
 \resizebox{\textwidth}{!}{%
 \begin{tabular}{lccccccccccc}
 \toprule
 \textbf{Variant}
 & \multicolumn{7}{c}{\textbf{Tensor components (PSNR\,$\uparrow$ / LEM\,$\downarrow$)}}
 & \multicolumn{4}{c}{\textbf{Parameter maps (PSNR\,$\uparrow$)}} \\
 \cmidrule(lr){2-8}
 \cmidrule(lr){9-12}
 & $D_{xx}$ & $D_{yy}$ & $D_{zz}$
 & $D_{xy}$ & $D_{xz}$ & $D_{yz}$ & LEM
 & MD & RD & FA & Color FA \\
 \midrule

 TensorLDM (Ours)
 & \textbf{36.45}$_{\text{\tiny 0.52}}$
 & \textbf{36.58}$_{\text{\tiny 0.66}}$
 & \textbf{36.75}$_{\text{\tiny 0.59}}$
 & 29.30$_{\text{\tiny 1.20}}$
 & 29.79$_{\text{\tiny 1.61}}$
 & 30.11$_{\text{\tiny 1.67}}$
 & \textbf{0.33}$_{\text{\tiny 0.02}}$
 & 32.86$_{\text{\tiny 0.82}}$
 & \textbf{33.15}$_{\text{\tiny 0.96}}$
 & \textbf{23.07}$_{\text{\tiny 0.44}}$
 & \textbf{25.03}$_{\text{\tiny 0.60}}$ \\
 \midrule
 \multicolumn{12}{l}{\textit{(a) Architecture \& geometric constraint}} \\
 \quad w/o CCA
 & 36.42$_{\text{\tiny 0.52}}$
 & 36.54$_{\text{\tiny 0.66}}$
 & 36.71$_{\text{\tiny 0.59}}$
 & 29.26$_{\text{\tiny 1.20}}$
 & 29.80$_{\text{\tiny 1.62}}$
 & 30.13$_{\text{\tiny 1.73}}$
 & 0.34$_{\text{\tiny 0.02}}$
 & 32.79$_{\text{\tiny 0.83}}$
 & 33.09$_{\text{\tiny 0.96}}$
 & 23.06$_{\text{\tiny 0.44}}$
 & 25.02$_{\text{\tiny 0.60}}$ \\
 \quad w/o MoE
 & 36.42$_{\text{\tiny 0.52}}$
 & 36.54$_{\text{\tiny 0.66}}$
 & 36.72$_{\text{\tiny 0.59}}$
 & 29.28$_{\text{\tiny 1.20}}$
 & 29.87$_{\text{\tiny 1.56}}$
 & \textbf{30.20}$_{\text{\tiny 1.67}}$
 & 0.34$_{\text{\tiny 0.02}}$
 & 32.79$_{\text{\tiny 0.83}}$
 & 33.10$_{\text{\tiny 0.96}}$
 & \textbf{23.07}$_{\text{\tiny 0.44}}$
 & 25.02$_{\text{\tiny 0.60}}$ \\
 \quad w/o AE-$\mathcal{L}_{\mathrm{LEM}}$
 & 36.19$_{\text{\tiny 0.53}}$
 & 36.33$_{\text{\tiny 0.67}}$
 & 36.53$_{\text{\tiny 0.61}}$
 & 29.29$_{\text{\tiny 1.20}}$
 & \textbf{29.90}$_{\text{\tiny 1.60}}$
 & 30.16$_{\text{\tiny 1.67}}$
 & 0.35$_{\text{\tiny 0.02}}$
 & 32.38$_{\text{\tiny 0.86}}$
 & 32.71$_{\text{\tiny 1.00}}$
 & 22.97$_{\text{\tiny 0.44}}$
 & 24.91$_{\text{\tiny 0.60}}$ \\
 \midrule
 \multicolumn{12}{l}{\textit{(b) Multi-stage training}} \\
 \quad w/o full-vol.\ fine-tuning
 & 35.88$_{\text{\tiny 0.49}}$
 & 36.02$_{\text{\tiny 0.63}}$
 & 36.17$_{\text{\tiny 0.56}}$
 & 29.20$_{\text{\tiny 1.17}}$
 & 29.57$_{\text{\tiny 1.47}}$
 & 29.91$_{\text{\tiny 1.67}}$
 & \textbf{0.33}$_{\text{\tiny 0.02}}$
 & 32.51$_{\text{\tiny 0.76}}$
 & 32.85$_{\text{\tiny 0.87}}$
 & 22.79$_{\text{\tiny 0.45}}$
 & 24.69$_{\text{\tiny 0.59}}$ \\
 \quad w/o joint AE
 & 36.23$_{\text{\tiny 0.52}}$
 & 36.25$_{\text{\tiny 0.66}}$
 & 36.48$_{\text{\tiny 0.61}}$
 & \textbf{29.84}$_{\text{\tiny 1.20}}$
 & 29.66$_{\text{\tiny 1.74}}$
 & 29.85$_{\text{\tiny 1.70}}$
 & 0.35$_{\text{\tiny 0.02}}$
 & 32.35$_{\text{\tiny 0.85}}$
 & 32.68$_{\text{\tiny 0.98}}$
 & 22.93$_{\text{\tiny 0.45}}$
 & 24.93$_{\text{\tiny 0.59}}$ \\
 \quad w/o joint LDM
 & 36.06$_{\text{\tiny 0.51}}$
 & 36.30$_{\text{\tiny 0.63}}$
 & 36.25$_{\text{\tiny 0.57}}$
 & 27.03$_{\text{\tiny 1.19}}$
 & 29.07$_{\text{\tiny 1.57}}$
 & 29.17$_{\text{\tiny 1.80}}$
 & 0.34$_{\text{\tiny 0.02}}$
 & \textbf{33.01}$_{\text{\tiny 0.71}}$
 & 33.08$_{\text{\tiny 0.83}}$
 & 22.53$_{\text{\tiny 0.48}}$
 & 24.54$_{\text{\tiny 0.62}}$ \\
 \bottomrule
 \end{tabular}
 }
\end{table*}

\paragraph{Computational Efficiency.}
TensorLDM requires approximately 123\,s for full-volume inference on a single NVIDIA A6000, compared to 11\,s for SuperDTI, 24\,s for the vanilla LDM, and 593\,s for the 2D slice-based Diff-DTI (\cref{app:runtime}). TensorLDM remains nearly $5{\times}$ faster than Diff-DTI while providing volumetric spatial consistency.

\subsection{Ablation Studies}

We ablate (a) architectural modules and geometric constraints, and (b) multi-stage training. In the variant names below, AE denotes the autoencoder and LDM the latent diffusion model. Each variant removes a single component while keeping all other settings identical (\cref{tab:ablation_ae_params}).

\subsubsection{Architecture and Geometric Constraint}

We evaluate three variants:
\textbf{w/o CCA} disables CCA in Stage~5 only (Stage~3 CCA remains active), removing learnable inter-component information exchange during denoising.
\textbf{w/o MoE} replaces the MoE DWI conditioner's learned gating with uniform averaging of the four DWI expert encoders, removing component-adaptive conditioning.
\textbf{w/o AE-$\mathcal{L}_{\mathrm{LEM}}$} removes the Log-Euclidean loss from Stage~3, relaxing the soft SPD constraint.

\begin{figure}[t]
 \centering
 \includegraphics[width=\linewidth]{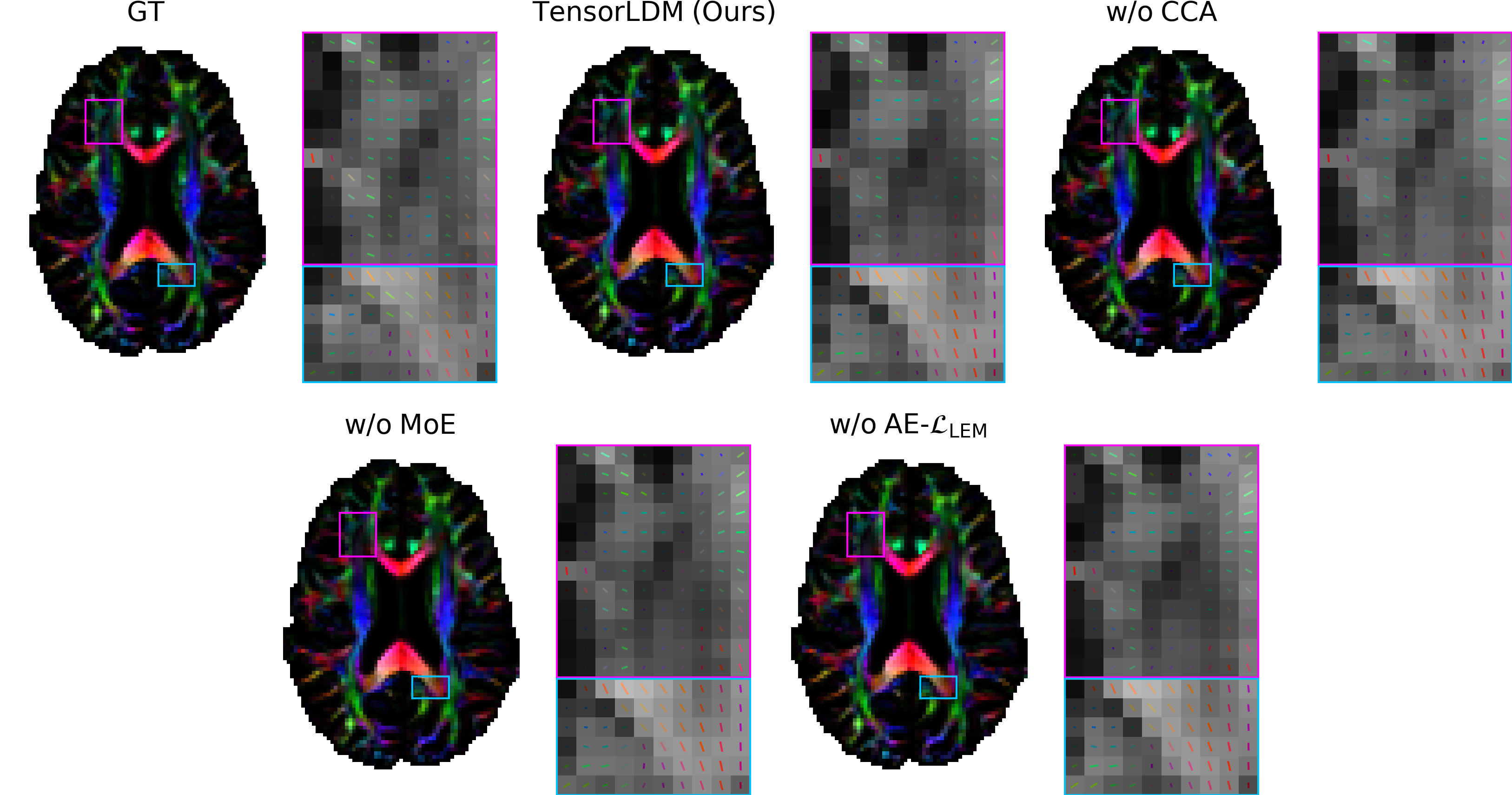}
 \caption{\textbf{Qualitative ablation results.} Architectural and geometric constraint variants. From left to right and top to bottom: ground truth (GT), TensorLDM (Ours), w/o CCA, w/o MoE, and w/o AE-$\mathcal{L}_{\mathrm{LEM}}$. Removing CCA or the MoE DWI conditioner leads to noisier FA maps and less coherent principal eigenvector ($\mathbf{v}_1$) fields, as highlighted in the magnified insets.}
 \label{fig:ablation_pretrain}
\end{figure}

Removing the Log-Euclidean loss (w/o AE-$\mathcal{L}_{\mathrm{LEM}}$) causes the largest degradation among architectural variants (LEM: 0.33 $\to$ 0.35), confirming that the geometric constraint is the most influential design choice. The MoE conditioner produces no significant change in whole-brain LEM (w/o MoE $p{=}0.177$, \cref{tab:ttest_ablation_summary}). The contribution of CCA and MoE is primarily to \emph{orientation coherence}: removing either degrades the principal-eigenvector field (\cref{fig:ablation_pretrain}). We detail these non-significant differences in~\cref{app:additional_results}.

\subsubsection{Impact of Multi-Stage Training}

We evaluate three training variants:
\textbf{w/o full-vol.\ fine-tuning} skips Stages~2--3, retaining only patch-based pre-training in Phase~I.
\textbf{w/o joint AE} skips Stage~3, removing CCA-based cross-component autoencoder refinement.
\textbf{w/o joint LDM} skips Stage~5, removing CCA during generation.

Full-volume fine-tuning contributes the largest gains for diagonal PSNR, while joint AE supports inter-component consistency (its removal yields one of the largest LEM increases, to 0.35). The w/o joint LDM variant achieves the highest MD PSNR but the lowest FA PSNR, indicating that CCA during denoising contributes to anisotropy and principal-eigenvector reconstruction rather than whole-brain LEM.

% Limitations moved to supplementary (app:limitations); brief mention in Conclusion.

\section{Conclusion}
\label{sec:conclusion}

We present TensorLDM, a component-wise latent diffusion framework for reconstructing diffusion tensors from critically undersampled DWIs. In the studied single-shell, four-volume (three-direction) setting, evaluation at the voxel, geometric, and functional levels shows that its advantage over strong baselines lies in the tensor geometry governing fiber tracking rather than per-voxel intensity, yielding the most accurate downstream tractography and near-ground-truth physical validity. Future work will target clinical cohorts, multi-shell protocols, faster sampling, and robustness across fitting algorithms and acquisition sites.

\noindent\textbf{Limitations.}
Our study is confined to a single-institution, single-shell cohort of healthy young adults, relies on one tensor-fitting algorithm for reference labels, and uses a stochastic sampler that adds latency and uncharacterized output variability (\cref{app:limitations}).

{
    \small
    \bibliographystyle{ieeenat_fullname}
    \bibliography{main}
}

\clearpage
\appendix
\section*{Appendix}
\section{Diffusion Tensor Model Details}
\label{app:dti}

\subsection{The Stejskal-Tanner Equation Explained}
The Stejskal-Tanner equation provides the foundational model for DTI~\cite{stejskal1965spin}. The terms are defined as follows:
\begin{itemize}
    \item \textbf{$S_0$}: The signal intensity measured in a non-diffusion-weighted acquisition (a $b{=}0$ image), where the diffusion-sensitizing gradients are turned off.
    \item \textbf{$S(\mathbf{g})$}: The signal intensity measured when a diffusion-sensitizing magnetic field gradient is applied along the direction of the unit vector $\mathbf{g}$.
    \item \textbf{b-value}: A scalar value that encapsulates the strength and duration of the diffusion gradients. A higher b-value results in greater signal attenuation for diffusing water molecules.
\end{itemize}

\subsection{Tensor Estimation and Clinical Context}
To solve for the six unknown components of the symmetric tensor $\mathbf{D}$, the Stejskal-Tanner equation must be sampled with at least six non-collinear gradient directions ($\mathbf{g}$). In clinical and research practice, many more directions (often 30 to 90 or more) are acquired to improve the accuracy and robustness of the tensor fit, especially in noisy data~\cite{jones2013white}. This sampling requirement leads to the primary clinical challenge of DTI: long acquisition times, which increase patient discomfort and susceptibility to motion artifacts.

\subsection{Tensor-Derived Metrics}
The diffusion tensor $\mathbf{D}$ is rarely interpreted directly. Instead, it is diagonalized to yield three eigenvalues ($\lambda_1 \geq \lambda_2 \geq \lambda_3$) and their corresponding eigenvectors ($\mathbf{v}_1, \mathbf{v}_2, \mathbf{v}_3$). These represent the magnitude of diffusion in three orthogonal directions and the orientation of those directions, respectively. From these, key microstructural metrics are calculated~\cite{basser1994mr}:
\begin{itemize}
    \item \textbf{Mean Diffusivity (MD)}: The average of the eigenvalues, $MD = (\lambda_1 + \lambda_2 + \lambda_3) / 3$. It measures the overall magnitude of water diffusion in a voxel, independent of directionality.
    \item \textbf{Radial Diffusivity (RD)}: The average of the secondary and tertiary eigenvalues, $RD = (\lambda_2 + \lambda_3) / 2$. This metric quantifies diffusion perpendicular to the principal fiber direction and is often cited as a marker of myelin integrity.
    \item \textbf{Fractional Anisotropy (FA)}: A normalized measure of the variance of the eigenvalues, indicating the degree to which diffusion is directional. An FA of 0 implies isotropic diffusion, while an FA close to 1 implies that diffusion is highly restricted to a single direction.
    \item \textbf{Color Fractional Anisotropy (Color FA)}: To visualize the principal fiber orientation alongside anisotropy, we modulate the FA map by the principal eigenvector $\mathbf{v}_1 = [v_{1x}, v_{1y}, v_{1z}]$. The resulting RGB image assigns colors based on direction: Red for left-right ($|v_{1x}|$), Green for anterior-posterior ($|v_{1y}|$), and Blue for superior-inferior ($|v_{1z}|$), scaled by the FA value.
\end{itemize}
These metrics are essential for the quantitative analysis of white matter integrity.

\section{Data and Preprocessing Details}
\label{app:data}

\textbf{Ground-Truth Tensor Generation:} The ground-truth DTI metrics for each subject were derived from the complete diffusion dataset, which included 18 $b{=}0$ volumes and 90 DWI volumes at $b{=}1000$~s/mm$^2$. Diffusion tensor fitting was performed using an ordinary linear least-squares method via the \texttt{dtifit} function in FSL~\cite{jenkinson2012fsl}, incorporating the provided gradient nonlinearity correction files. This process yielded the full diffusion tensor, from which all ground-truth metrics, including fractional anisotropy (FA) and mean diffusivity (MD), were calculated~\cite{glasser2013minimal}.

\textbf{Undersampled Input Selection:} The 4-volume sparse input for our model was created by selecting a specific subset of DWIs. For the $b{=}1000$~s/mm$^2$ shell, we identified the three diffusion gradient vectors most closely aligned with the standard Cartesian axes ([1, 0, 0], [0, 1, 0], and [0, 0, 1]) by minimizing the Euclidean distance. The corresponding DWI volumes were extracted, and a single $b{=}0$ volume was prepended to form the final 4-volume input stack $\mathbf{X} = \{\mathbf{X}_c\}_{c=0}^{3}$.

\section{Implementation Details}
\label{app:impl}

\subsection{Training Pipeline and Phase-Wise Details}

\paragraph{Overall Pipeline.}
TensorLDM is trained in a two-phase pipeline comprising five sequential stages (three in Phase~I and two in Phase~II). Phase~I (Stages~1--3) establishes the high-fidelity component-wise latent space via the Anatomy-Conditioned Autoencoder, while Phase~II (Stages~4--5) trains the component-wise latent diffusion model within that space.

\paragraph{Phase~I: Component-Wise Anatomy-Conditioned Autoencoder.}
\textbf{Stage~1: Local Component Learning (Patch-based).} The base autoencoder is first trained on local patches ($32{\times}32{\times}32$) to learn a high-fidelity latent representation for each of the six tensor components independently. This stage focuses on capturing fine-grained structural details and local diffusion patterns. We use the Adam optimizer with a learning rate of $1.0\times10^{-5}$ and a batch size of 2. The objective combines voxel-wise reconstruction loss and a vector-quantization commitment loss~\cite{van2017neural}.

\textbf{Stage~2: Global Structure Fine-tuning (Full-volume).} Following the patch-based pre-training, the autoencoder is further fine-tuned on full-volume data. This transition from local to global learning (curriculum learning) ensures global spatial consistency across the latent field and stabilizes the representation for downstream tasks. The learning rate is maintained at $1.0\times10^{-5}$.

\textbf{Stage~3: Cross-Component Joint Refinement.} After establishing the base latent space, we freeze the encoder and conditioner weights and fine-tune the Cross-Component Attention blocks, the decoder output layers, and a lightweight joint fusion MLP. The fusion MLP receives all six component latents and promotes cross-component consistency via cross-component attention. Optimization uses the Adam optimizer with a learning rate of $1.0\times10^{-5}$ and an effective batch size of 16. To encourage symmetric positive definiteness (SPD) and geometric validity, the loss function is augmented with a Log-Euclidean metric term ($\lambda_{\mathrm{LEM}} = 0.01$).

\paragraph{Phase~II: Component-Wise Latent Diffusion.}
\textbf{Stage~4: Independent Component Pre-training.} We pre-train the denoising U-Net on individual tensor components ($d \in \{0,\ldots,5\}$) with Cross-Component Attention deactivated. Critically, the model is \emph{conditioned on DWI features} via the MoE DWI conditioner, which takes multi-scale feature maps extracted by the shared conditioner $\mathcal{C}$ from each DWI volume ($b{=}0, b_{1000,x}, b_{1000,y}, b_{1000,z}$), processes them through per-volume expert encoders, and computes component-adaptive gating weights to selectively combine the expert outputs. This stage learns the mapping from sparse DWI signals to individual tensor components without modeling inter-component dependencies. The diffusion process uses a linear beta schedule~\cite{ho2020denoising} over 1000 steps. Training uses the AdamW optimizer with a learning rate of $1.0\times10^{-5}$ and a batch size of 3 (effective batch size of 12 with 4$\times$ gradient accumulation).

\textbf{Stage~5: Joint Component Fine-tuning.} Building on Stage~4, we activate the Cross-Component Attention (CCA) blocks to enable information exchange between all six tensor components during denoising. The model continues to use component-adaptive conditioning via the MoE DWI conditioner while now promoting cross-component consistency through joint processing. This encourages compliance with physical constraints (symmetric positive definiteness) and promotes anatomically coherent directional patterns. Training uses AdamW with a learning rate of $1.0\times10^{-5}$, initialized from the Stage~4 checkpoint.

\subsection{Detailed Network Configurations}

\paragraph{Joint MLP Block Architecture.}
The joint refinement block added at the end of the decoder operates voxel-wise across all six tensor components. Its architecture is as follows:

\begin{itemize}
\item \textbf{Input dimension}: 1536
(= 256 channels $\times$ 6 tensor components)
\item \textbf{Hidden dimension}: 768 (GELU activation)
\item \textbf{Output dimension}: 1536
\item \textbf{Total parameters}: 2{,}361{,}600
\end{itemize}

This block contributes only a small fraction of the decoder's total parameters and is designed to promote cross-component consistency with minimal overhead.

\subsection{Ablation Training and Analysis Details}

\paragraph{Training Schedules for Ablation Variants.}
For completeness, we report the exact number of training iterations used for each ablation model, ensuring that performance differences are attributable to architectural factors rather than computational budget:
\begin{itemize}
\item \textbf{TensorLDM (full model)}:
75k steps (autoencoder pre-training; Stages~1--2) + 5k (joint refinement)
+ 25k (independent component pre-training) + 25k (joint component fine-tuning)
\item \textbf{No-Anatomy}: 120k autoencoder-only steps
\item \textbf{No-Pretrain}: 100k diffusion steps
\item \textbf{Channel-Shared variant}: 20k diffusion steps
\end{itemize}
These figures report the training budget for each variant; all are trained to convergence so that observed differences reflect architecture rather than budget (for example, No-Pretrain matches the w/o joint LDM ablation in the main paper).

\paragraph{Effect of Patch Pre-training (Stage 1).}
At matched compute, we compare the full Stage~1+2 curriculum against a Stage~2-only variant on the $n{=}195$ HCP test subjects. Stage~1 deliberately trades a small amount of whole-brain mean fidelity for eigenstructure accuracy: whole-brain LEM is 0.520 \vs\ 0.508 (Stage~2-only better; $\Delta{=}{+}0.012$, paired Wilcoxon $p{=}6.6\times10^{-11}$), whereas off-diagonal LEM is 0.240 \vs\ 0.255 (Stage~1+2 better; $\Delta{=}{-}0.015$, $p{=}1.3\times10^{-25}$). Because diagonal magnitudes dominate whole-brain LEM while off-diagonal errors propagate into eigenvector orientation (over which streamline tractography integrates), this trade-off favors the tensor geometry that matters for downstream fiber tracking.

\section{Competing Methods Details}
\label{app:baselines}
Unless otherwise noted, all learning-based baselines are implemented as 3D networks and trained on full 3D volumes to ensure a rigorous comparison with our volumetric approach.

\begin{itemize}
    \item \textbf{ADE (Analytic Diagonal Estimation):} A non-learning baseline that assumes a diagonal diffusion tensor. The diagonal components ($D_{xx}$, $D_{yy}$, $D_{zz}$) are computed from the log-linearized Stejskal-Tanner equation using the most aligned gradients. Off-diagonal elements are set to zero, and the tensor is projected onto the symmetric positive definite (SPD) manifold.

    \item \textbf{SuperDTI~\cite{li2021superdti}:} Implemented using a 2D U-Net-style encoder--decoder CNN with residual learning. Following the original protocol, the network learns the nonlinear mapping from uniformly sampled sparse DWI signals directly to FA, MD, and the primary eigenvectors (or color maps) using an $\ell_2$ regression loss. We use the architecture specified in the original paper, adapted for 2D-based processing.

    \item \textbf{Diff-DTI~\cite{zhang2024diff}:} A conditional score-based diffusion model. The method operates on 2D slices to directly synthesize DTI parameter maps (\eg, FA, MD, Color FA) from a few conditional DWIs. The model uses a U-Net backbone enhanced by a Feature Enhancement Fusion Mechanism (FEFM), which integrates a Transformer-based auxiliary path to preserve fine structural details, and is trained with a score-matching objective.

    \item \textbf{CycleGAN~\cite{zhu2017unpaired}:} The architecture consists of two 3D U-Net generators and two 3D PatchGAN discriminators, trained with an adversarial loss and an $\ell_1$ cycle-consistency loss ($\lambda=10$).

    \item \textbf{Pix2Pix~\cite{isola2017image}:} The generator is a 3D U-Net, and the discriminator is a 3D PatchGAN. The training objective is a sum of a vanilla GAN loss and an $\ell_1$ reconstruction loss ($\lambda_{\ell_1} = 100$).

    \item \textbf{ResViT~\cite{dalmaz2022resvit}:} ResViT employs a hybrid architecture combining a 3D ResNet-style backbone with interleaved Vision Transformer blocks to capture long-range dependencies. It is trained with a composite $\ell_1$ and adversarial loss.

    \item \textbf{LDM~\cite{rombach2022high}:} A standard Latent Diffusion Model serves as a baseline.
We employ a conventional latent autoencoder \emph{without DWI-aided decoding or anatomical feature injection}, thereby reproducing the canonical LDM setup.
The diffusion U-Net is conditioned solely through channel-wise concatenation of the DWI latent and the noisy tensor latent, following the design principles of SR3~\cite{saharia2022image} and Palette~\cite{saharia2022palette}.
This configuration reflects the standard conditional LDM mechanism and stands in contrast to the component-wise framework and shared DWI conditioning used in TensorLDM.
\end{itemize}

\subsection{Reconstructing Diffusion Tensors from Parameter Maps}
\label{app:recon_from_params}

SuperDTI and Diff-DTI do not directly output the full diffusion tensor, but instead predict
diffusion parameter maps such as mean diffusivity (MD), radial diffusivity (RD), fractional anisotropy (FA),
and Color FA (RGB-encoded FA).
To enable tensor-domain metrics and tractography, we reconstruct an approximate diffusion tensor
$\mathbf{D} \in \mathbb{R}^{3{\times}3}$ in each voxel from these quantities under a cylindrically symmetric model.

We assume a single-shell acquisition and impose
\[
\lambda_2 = \lambda_3 = \mathrm{RD},
\]
so that the tensor has one principal eigenvalue $\lambda_1 = \mathrm{AD}$ (axial diffusivity)
and two identical secondary eigenvalues $\lambda_2 = \lambda_3 = \mathrm{RD}$.
Using the standard relation
\[
\mathrm{MD} = \frac{\lambda_1 + \lambda_2 + \lambda_3}{3}
          = \frac{\mathrm{AD} + 2\,\mathrm{RD}}{3},
\]
we recover axial diffusivity as
\[
\mathrm{AD} = 3\,\mathrm{MD} - 2\,\mathrm{RD}.
\]
In practice, we clip negative values of AD to zero to avoid clearly unphysical eigenvalues.

The principal eigenvector $\mathbf{v}_1 \in \mathbb{R}^3$ is estimated from the Color FA image.
Let $\mathbf{f}_{\mathrm{CFA}} = (f_x, f_y, f_z)$ denote the Color FA vector at a voxel;
we normalize it to obtain
\[
\mathbf{v}_1 = \frac{\mathbf{f}_{\mathrm{CFA}}}{\lVert\mathbf{f}_{\mathrm{CFA}}\rVert_2 + \varepsilon},
\]
with a small $\varepsilon > 0$ for numerical stability.
To suppress unreliable orientations in nearly isotropic voxels, we set $\mathbf{v}_1 = \mathbf{0}$ whenever
$\mathrm{FA} < 0.05$.

Under these assumptions, the reconstructed diffusion tensor is given by the rank-1 update
\[
\mathbf{D} = \mathrm{RD}\,\mathbf{I}_3 + (\mathrm{AD} - \mathrm{RD})\, \mathbf{v}_1 \mathbf{v}_1^\top,
\]
where $\mathbf{I}_3$ is the $3{\times}3$ identity matrix.
Writing $\mathbf{v}_1 = (v_x, v_y, v_z)$ and $\Delta = \mathrm{AD} - \mathrm{RD}$, the six independent
components are
\begin{align*}
D_{xx} &= \mathrm{RD} + \Delta\, v_x^2, \\
D_{yy} &= \mathrm{RD} + \Delta\, v_y^2, \\
D_{zz} &= \mathrm{RD} + \Delta\, v_z^2, \\
D_{xy} &= \Delta\, v_x v_y, \\
D_{xz} &= \Delta\, v_x v_z, \\
D_{yz} &= \Delta\, v_y v_z.
\end{align*}
We then store $(D_{xx}, D_{yy}, D_{zz}, D_{xy}, D_{xz}, D_{yz})$ as a 4D volume
and use this reconstructed tensor field for all tensor-domain metrics (\eg, LEM)
and tractography-based evaluations reported for SuperDTI and Diff-DTI.

\textbf{On comparison fairness.} Because SuperDTI and Diff-DTI regress scalar maps, off-diagonal tensor structure is simply not part of their output. The cylindrically symmetric reconstruction above is the most faithful tensor consistent with the quantities they do predict. The resulting lower off-diagonal and LEM scores therefore characterize the scalar-map design of these methods, not an artifact introduced by our evaluation. We deliberately retain their original 2D formulations (rather than re-engineering them into 3D six-component tensor regressors) so that the benchmark measures the published methods and exposes precisely the architectural limitation (loss of directional/off-diagonal information) that motivates direct full-tensor reconstruction. Retraining them into a different output space would create new models rather than evaluate the existing ones.

\section{Diffusion Tensor Metrics and Manifold Geometry}
\label{app:metrics}

To provide a geometrically rigorous assessment of the reconstructed diffusion tensors, we employ the Log-Euclidean Metric (LEM) in addition to conventional image-based metrics (PSNR, SSIM). This addresses the fundamental limitation that the space of diffusion tensors does not conform to Euclidean geometry.

\subsection{Limitation of Euclidean Metrics}

A diffusion tensor $\mathbf{D}$ is represented by a $3{\times}3$ SPD matrix. The space of all such matrices, $\mathcal{S}^{3}_{++}$, forms a nonlinear Riemannian manifold rather than a flat Euclidean space. Applying standard Euclidean metrics (\eg, $\ell_2$ norm, PSNR, SSIM) to the six unique tensor components treats them as a simple 6D vector. This approach ignores:
\begin{itemize}
    \item \textbf{Physical Plausibility}: It does not enforce the positive definiteness of the tensor (\ie, strictly positive eigenvalues), which is a core physical constraint of the diffusion tensor.
    \item \textbf{Geodesic Distance}: The resulting distance metric does not correspond to the shortest path (geodesic) between two tensors on the manifold, leading to potentially inaccurate assessments in regions of high anisotropy or complex fiber geometry.
\end{itemize}

\subsection{Log-Euclidean Metric (LEM)}

The Log-Euclidean Metric (LEM)~\cite{arsigny2006log} provides an effective and computationally tractable distance metric for SPD matrices. It utilizes the matrix logarithm ($\mathrm{Log}$) to map the curved SPD manifold ($\mathcal{S}^{3}_{++}$) to a flat vector space (the space of symmetric matrices, $\mathcal{S}^{3}$), where standard Euclidean operations become valid.

The Log-Euclidean distance between two tensors, $\mathbf{D}_1$ and $\mathbf{D}_2$, is defined as the Frobenius norm of the difference between their matrix logarithms:
\[
d_{\mathrm{LE}}(\mathbf{D}_1, \mathbf{D}_2) = \lVert\mathrm{Log}(\mathbf{D}_1) - \mathrm{Log}(\mathbf{D}_2)\rVert_F.
\]
Here, $\mathrm{Log}(\mathbf{D})$ is computed via eigendecomposition ($\mathbf{D} = \mathbf{V} \boldsymbol{\Lambda} \mathbf{V}^\top$, where $\boldsymbol{\Lambda}$ contains the eigenvalues $\lambda_i$), such that $\mathrm{Log}(\mathbf{D}) = \mathbf{V} \log(\boldsymbol{\Lambda}) \mathbf{V}^\top$. In our implementation, we ensure numerical stability by clamping all eigenvalues to a minimum positive value before applying the logarithm. By minimizing this metric, we encourage the reconstructed tensor to not only match the ground truth in component values but also adhere to the appropriate underlying tensor geometry, supporting physically plausible downstream analyses such as tractography.

\section{Tractography Evaluation: Tract Core Distance}
\label{app:tract_core_distance}

To assess how tensor reconstruction quality affects downstream fiber tracking (\cref{sec:experiments}), we measure the geometric discrepancy between white matter bundles obtained from reconstructed tensors and those from the reference tensors using a \emph{tract core distance}~\cite{garyfallidis2014dipy, girard2014towards}.

For each subject, whole-brain probabilistic tractography is run on the reference and on each reconstructed tensor field with identical MRtrix3 seeding and tracking parameters~\cite{tournier2019mrtrix3}. Major bundles (\eg, corticospinal tract, cingulum, corpus callosum segments) are segmented from the whole-brain tractograms using TractSeg-derived bundle masks. For each bundle, we build a smooth 3D \emph{core} trajectory that summarizes its geometry: all streamlines are resampled to a fixed number of points in world coordinates, and a low-degree polynomial is fitted to the cross-sectional centroid positions along the main bundle direction. This yields a core trajectory for the ground-truth tensors, $\gamma^{\text{GT}}$, and for each reconstruction, $\gamma^{\text{rec}}$.

The tract core distance for a bundle is defined as the average nearest-point distance from the reconstructed core to the reference core:
\[
d_{\text{core}}(\gamma^{\text{GT}}, \gamma^{\text{rec}})
= \frac{1}{M} \sum_{m=1}^{M}
\min_{1 \le k \le K}
\big\lVert \gamma^{\text{GT}}(k) - \gamma^{\text{rec}}(m) \big\rVert_2,
\]
where $\{\gamma^{\text{GT}}(k)\}_{k=1}^{K}$ and $\{\gamma^{\text{rec}}(m)\}_{m=1}^{M}$ are the sampled points along the two core trajectories in physical space (mm). For 2\,mm isotropic data, we treat sub-voxel discrepancies as negligible by thresholding very small distances. Lower values of $d_{\text{core}}$ indicate that the reconstructed tensor yields a bundle whose central trajectory closely matches that of the ground truth.

\section{Additional Qualitative and Quantitative Results}
\label{app:additional_results}

\begin{figure*}[t]
    \centering
    \includegraphics[width=\textwidth]{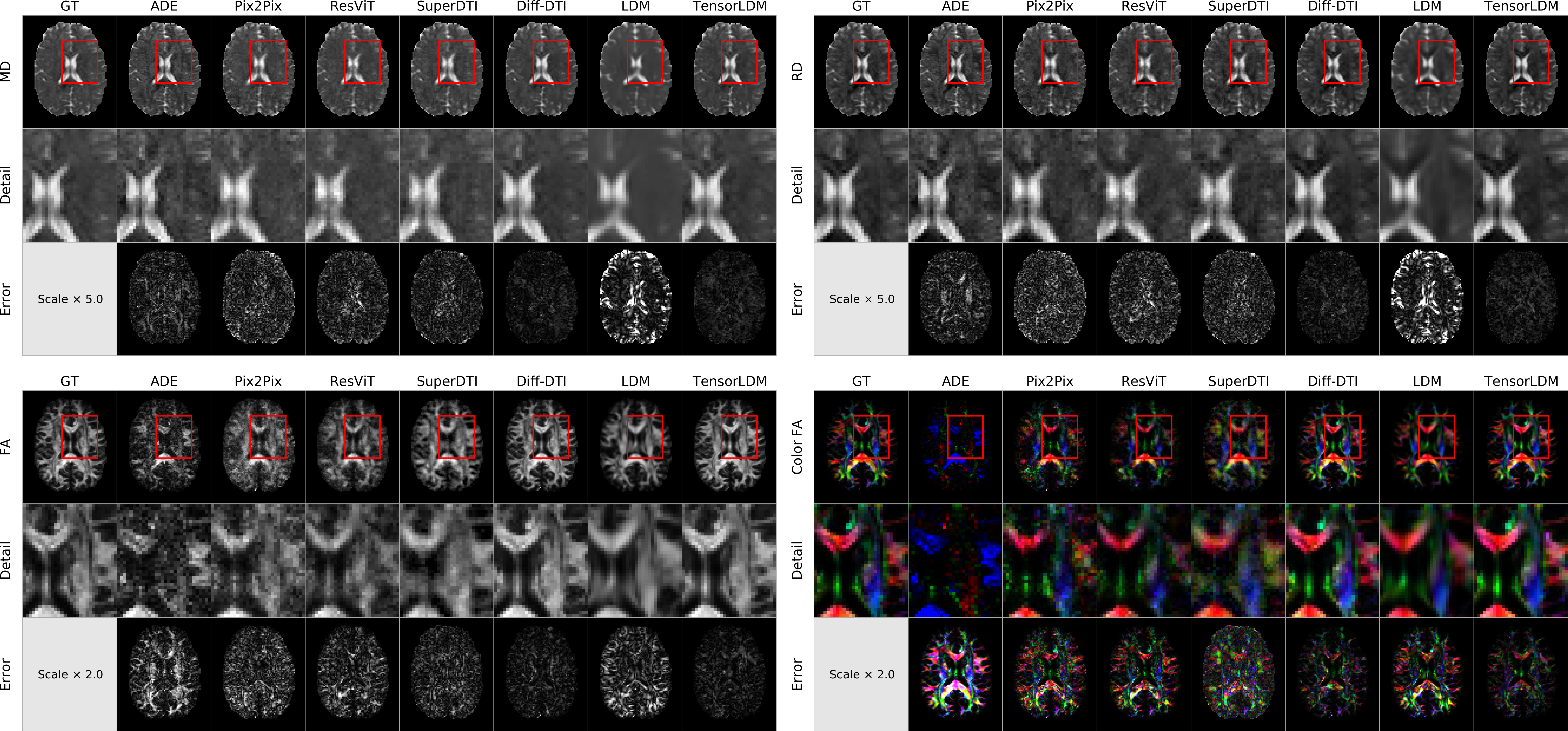}
    \caption{\textbf{Qualitative comparison of DTI parameter maps on an axial slice.} Visualization of MD, RD, FA, and Color FA for the same subject shown in the main text. TensorLDM consistently provides a faithful reconstruction with reduced noise compared to baselines.}
    \label{fig:qual_params}
\end{figure*}

\begin{figure*}[t]
    \centering
    \includegraphics[width=\textwidth]{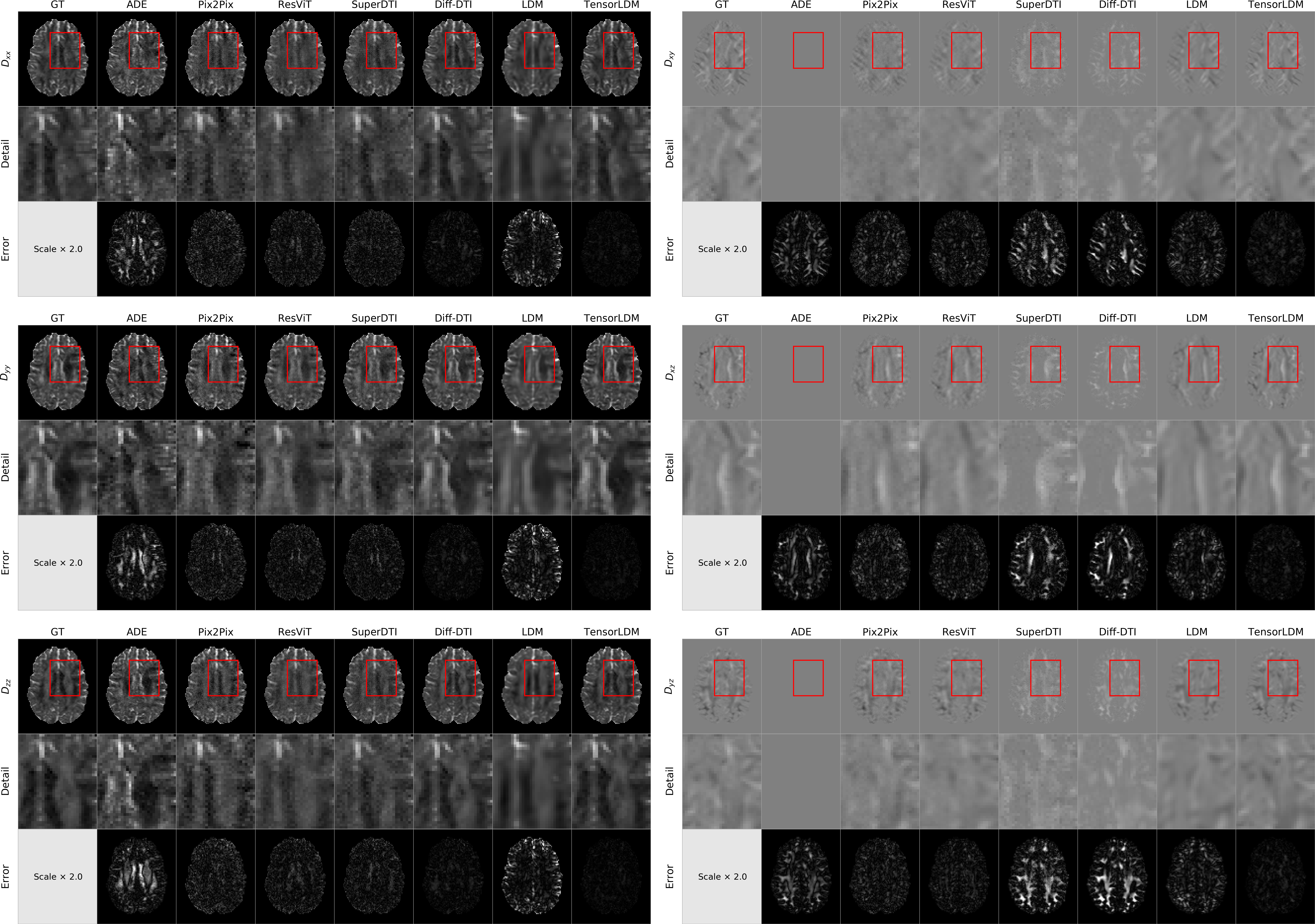}
    \caption{\textbf{Qualitative comparison of diffusion tensor components on an axial slice.} Visualization of the six individual tensor components for the same subject. TensorLDM provides a faithful reconstruction across all components, with reduced noise and artifacts, particularly in the off-diagonal elements.}
    \label{fig:qual_tensors}
\end{figure*}

\begin{figure*}[t]
    \centering
    \includegraphics[width=\textwidth]{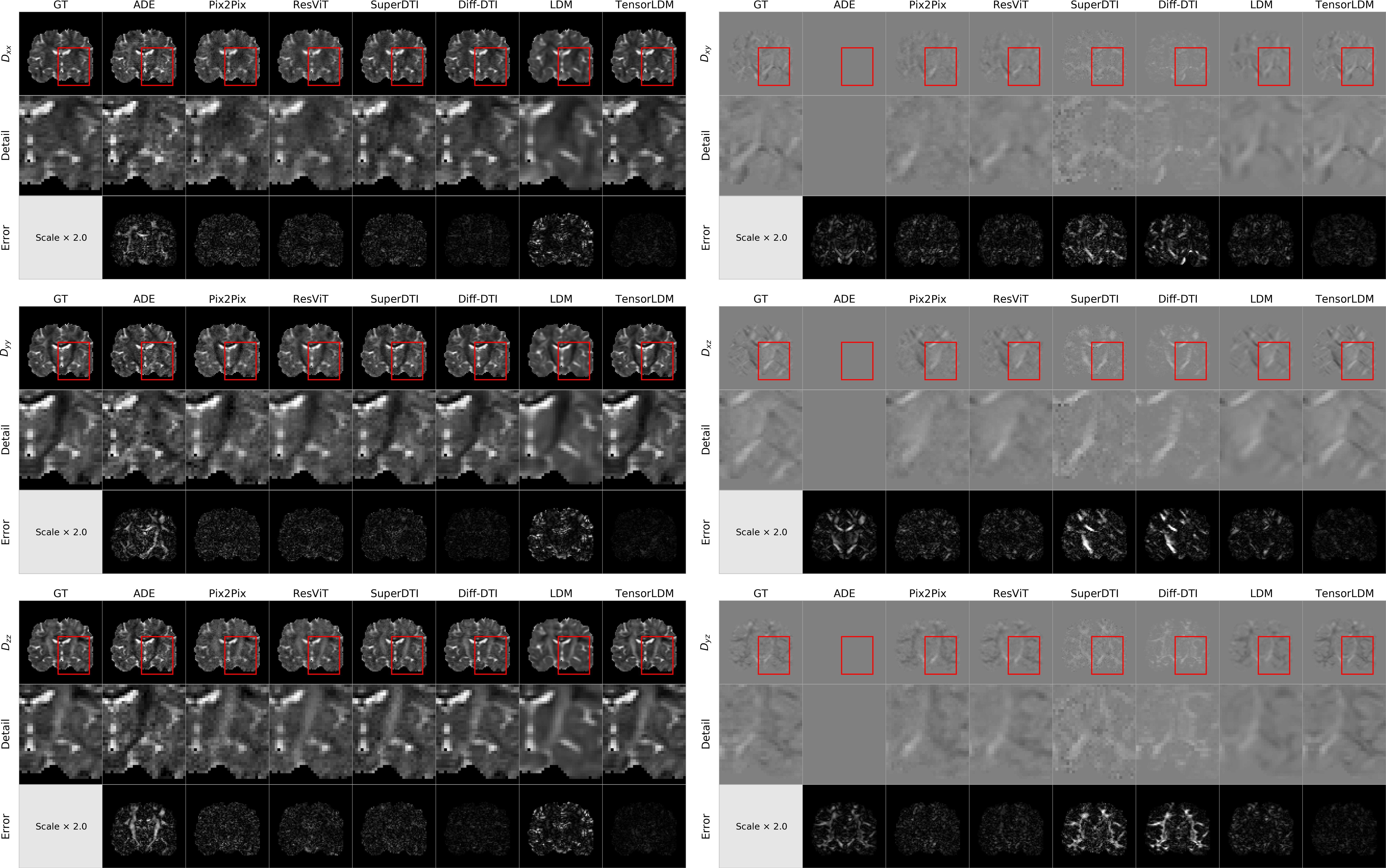}
    \caption{\textbf{Qualitative comparison of diffusion tensor components on a coronal slice.} Reconstructed components, magnified detail, and error maps on a coronal slice. TensorLDM provides the most faithful reconstruction, particularly for the challenging off-diagonal components.}
    \label{fig:qual_tensors_coronal}
\end{figure*}

This section contains additional results, including axial views of the DTI parameter maps (\cref{fig:qual_params}) and tensor components (\cref{fig:qual_tensors}), a coronal view of the tensor components (\cref{fig:qual_tensors_coronal}), detailed statistical significance tests (\cref{tab:ttest_summary,tab:ttest_tensors_summary,tab:ttest_ablation_summary}), and inference runtime comparisons (\cref{tab:runtime}).

\begin{table*}[t]
\caption{\textbf{Paired t-test results (TensorLDM \vs\ baselines) for DTI parameter maps.}
Each entry tests whether TensorLDM is significantly better (lower NMSE, higher PSNR/SSIM) than the compared method via two-sided paired t-test.
Legend: *** $p<10^{-3}$, ** $p<10^{-2}$, * $p<0.05$, (n.s.) $p \ge 0.05$.
}
\label{tab:ttest_summary}
\centering
\scriptsize
\setlength{\tabcolsep}{4pt}

\resizebox{\textwidth}{!}{%
\begin{tabular}{llccccccc}
\toprule
\textbf{Metric} & \textbf{Measure}
 & ADE & CycleGAN & Pix2Pix & ResViT & Diff-DTI & SuperDTI & LDM \\
\midrule
MD & NMSE
  & *** & *** & *** & *** & *** & *** & *** \\
FA  & NMSE
  & *** & *** & *** & *** & 0.097 (n.s.) & *** & *** \\
RD  & NMSE
  & *** & *** & *** & *** & *** & *** & *** \\
\midrule
MD & PSNR
  & *** & *** & *** & *** & *** & *** & *** \\
FA  & PSNR
  & *** & *** & *** & *** & *** & *** & *** \\
RD  & PSNR
  & *** & *** & *** & *** & *** & *** & *** \\
\midrule
MD & SSIM
  & *** & *** & *** & *** & *** & *** & *** \\
FA  & SSIM
  & *** & *** & *** & *** & *** & *** & *** \\
RD  & SSIM
  & *** & *** & *** & *** & *** & *** & *** \\
\bottomrule
\end{tabular}%
}
\end{table*}

\begin{table*}[t]
\caption{\textbf{Paired t-test results (TensorLDM \vs\ baselines) for tensor components.}
Each entry tests whether TensorLDM is significantly better (lower LEM, higher PSNR/SSIM) via two-sided paired t-test.
Legend: *** $p<10^{-3}$, ** $p<10^{-2}$, * $p<0.05$, (n.s.) $p \ge 0.05$.
}
\label{tab:ttest_tensors_summary}
\centering
\scriptsize
\setlength{\tabcolsep}{4pt}

\resizebox{\textwidth}{!}{%
\begin{tabular}{llccccccc}
\toprule
\textbf{Metric} & \textbf{Component / Measure}
 & ADE & CycleGAN & Pix2Pix & ResViT & Diff-DTI & SuperDTI & LDM \\
\midrule
LEM & --
 & *** & *** & *** & *** & *** & *** & *** \\
\midrule
PSNR & $D_{xx}$ & *** & *** & *** & *** & *** & *** & *** \\
PSNR & $D_{yy}$ & *** & *** & *** & *** & *** & *** & *** \\
PSNR & $D_{zz}$ & *** & *** & *** & *** & *** & *** & *** \\
PSNR & $D_{xy}$ & *** & *** & *** & *** & *** & *** & *** \\
PSNR & $D_{xz}$ & *** & *** & *** & *** & *** & *** & *** \\
PSNR & $D_{yz}$ & *** & *** & *** & *** & *** & *** & *** \\
\midrule
SSIM & $D_{xx}$ & *** & *** & *** & *** & *** & *** & *** \\
SSIM & $D_{yy}$ & *** & *** & *** & *** & *** & *** & *** \\
SSIM & $D_{zz}$ & *** & *** & *** & *** & *** & *** & *** \\
SSIM & $D_{xy}$ & *** & *** & *** & *** & *** & *** & *** \\
SSIM & $D_{xz}$ & *** & *** & *** & *** & *** & *** & *** \\
SSIM & $D_{yz}$ & *** & *** & *** & *** & *** & *** & *** \\
\bottomrule
\end{tabular}%
}
\end{table*}

\begin{table*}[t]
\caption{\textbf{Paired t-test results for ablation study.}
Each entry tests whether the full TensorLDM is significantly better than each ablated variant via two-sided paired t-test.
Legend: *** $p<10^{-3}$, ** $p<10^{-2}$, * $p<0.05$, (n.s.) $p\ge 0.05$.
}
\label{tab:ttest_ablation_summary}
\centering
\scriptsize
\setlength{\tabcolsep}{4pt}

\resizebox{\textwidth}{!}{%
\begin{tabular}{lcccccc}
\toprule
\textbf{Metric / Component}
 & w/o CCA & w/o MoE & w/o AE-$\mathcal{L}_{\mathrm{LEM}}$ & w/o full-vol.\ fine-tuning & w/o joint AE & w/o joint LDM \\
\midrule
LEM & ** & 0.177 (n.s.) & *** & *** & *** & 0.075 (n.s.) \\
MD--PSNR & *** & *** & *** & *** & *** & *** \\
FA--PSNR  & 0.247 (n.s.) & 0.682 (n.s.) & *** & *** & *** & *** \\
RD--PSNR  & *** & *** & *** & *** & *** & 0.124 (n.s.) \\
\cmidrule(lr){1-7}
$D_{xx}$ & *** & *** & *** & *** & *** & *** \\
$D_{yy}$ & *** & *** & *** & *** & *** & *** \\
$D_{zz}$ & *** & *** & *** & *** & *** & *** \\
$D_{xy}$ & *** & *** & *** & *** & *** & *** \\
$D_{xz}$ & 0.927 (n.s.) & 0.164 (n.s.) & 0.055 (n.s.) & *** & 0.070 (n.s.) & *** \\
$D_{yz}$ & 0.808 (n.s.) & 0.195 (n.s.) & 0.543 (n.s.) & ** & ** & *** \\
\bottomrule
\end{tabular}%
}
\end{table*}

\textbf{Discussion of non-significant LEM differences.} Two whole-brain LEM comparisons in \cref{tab:ttest_ablation_summary} do not reach significance: w/o MoE ($p{=}0.177$) and w/o joint LDM ($p{=}0.075$). We report this directly rather than omit it. Whole-brain LEM is dominated by the larger diagonal magnitudes ($D_{xx}, D_{yy}, D_{zz}$), so modules that chiefly refine off-diagonal and orientation structure shift the global scalar only slightly. The same table makes this concrete: removing the joint diffusion stage (w/o joint LDM) significantly degrades all three off-diagonal components ($D_{xy}, D_{xz}, D_{yz}$, $p<10^{-3}$) and FA-PSNR while leaving whole-brain LEM unchanged, and removing CCA or MoE leaves several off-diagonal/FA comparisons non-significant even where diagonal PSNR changes significantly. The qualitative ablation (\cref{fig:ablation_pretrain}) shows that removing CCA or the MoE conditioner visibly disrupts the principal-eigenvector ($\mathbf{v}_1$) field. We therefore position CCA, the MoE conditioner, and the joint diffusion stage as improving principal-eigenvector alignment and local tensor structure, while the Log-Euclidean autoencoder constraint (w/o AE-$\mathcal{L}_{\mathrm{LEM}}$) is the component that most affects whole-brain LEM.

\section{Inference Runtime Analysis}
\label{app:runtime}

\Cref{tab:runtime} summarizes the end-to-end inference time for a full 2\,mm whole-brain volume on a single NVIDIA A6000. TensorLDM incurs additional computational cost relative to a vanilla LDM due to Cross-Component Attention and tensor-aware refinement, while achieving markedly higher tensor consistency. Importantly, TensorLDM remains substantially faster than 2D slice-based diffusion models such as Diff-DTI, which require processing each slice independently and accumulate significant overhead.

The iterative diffusion sampler dominates TensorLDM's cost. The cross-method figures in \cref{tab:runtime} were measured on an NVIDIA A6000. On a common NVIDIA RTX 6000 Ada, end-to-end inference is $32.7 \pm 0.2$ s/vol with the full 250-step DDIM sampler and $9.4 \pm 0.0$ s/vol with 25 DDIM steps (encode 4.3 s, sample $25.8 \rightarrow 2.6$ s, decode 2.3 s; mean over $n{=}5$ volumes), \ie a $\sim$3.5$\times$ reduction on the same device while leaving the reconstruction qualitatively unchanged.

\begin{table}[H]
\caption{\textbf{Inference runtime on a single NVIDIA A6000~(48\,GB).}
TensorLDM is slower than a vanilla LDM due to Cross-Component Attention and tensor-aware refinement,
yet remains substantially faster than 2D slice-based diffusion methods.
}
\label{tab:runtime}
\centering
\scriptsize
\begin{tabular}{lc}
\toprule
\textbf{Model} & \textbf{Inference Time (s)} \\
\midrule
SuperDTI & 10.93 \\
LDM & 23.58 \\
TensorLDM (Ours) & 122.96 \\
Diff-DTI & 593.22 \\
\bottomrule
\end{tabular}
\end{table}

\section{SPD Violation Analysis}
\label{app:spd_violation}

\Cref{tab:spd_violation} reports the percentage of brain voxels where the reconstructed diffusion tensor violates the SPD constraint (\ie, at least one eigenvalue is negative). Eigenvalues are computed via symmetric eigendecomposition of the $3{\times}3$ tensor assembled from the six predicted components at each voxel within the brain mask. The ground-truth tensors themselves, obtained from 90-direction linear least-squares fitting, exhibit 1.40\% SPD violations, as ordinary least-squares does not enforce positive definiteness.

TensorLDM achieves a violation rate comparable to the ground truth, and removing the Log-Euclidean loss (w/o AE-$\mathcal{L}_{\mathrm{LEM}}$) increases it, consistent with this soft geometric constraint contributing to SPD preservation (\cref{tab:spd_violation}).

\begin{table}[H]
\caption{\textbf{SPD violation rates across methods.} Values report the mean percentage ($\pm$ std) of brain voxels with at least one negative eigenvalue, computed over 195 test subjects.}
\label{tab:spd_violation}
\centering
\scriptsize
\begin{tabular}{lc}
\toprule
\textbf{Method} & \textbf{SPD Violation (\%)} \\
\midrule
GT (90-dir) & $1.40 \pm 0.27$ \\
\midrule
TensorLDM (Ours) & $1.54 \pm 0.28$ \\
\quad w/o AE-$\mathcal{L}_{\mathrm{LEM}}$ & $1.61 \pm 0.29$ \\
\bottomrule
\end{tabular}
\end{table}

\section{Latent Representation Probing}
\label{app:probe}

To test whether the Anatomy-Conditioned Autoencoder encourages a division of labor between anatomical structure (carried by the DWI conditioning) and tensor properties (carried by the latents), we freeze the trained model and fit a linear probe, evaluated subject-wise, that predicts voxel-wise fractional anisotropy (FA) from (i) the anatomical conditioning representation $\Phi_{\mathbf{X}}$ and (ii) the tensor-component latents $\{\mathbf{z}^d_{\mathbf{Y}}\}$. FA is largely recoverable from the anatomical representation ($R^2 = 0.61$) but is not linearly carried by the tensor latents ($R^2 = 0.08$), consistent with the two encoding complementary rather than redundant information. This is a single linear probe on one scalar (FA) and does not by itself establish formal disentanglement; it is computed on the frozen model and requires no retraining.

\section{Component-Wise Architecture Details}
\label{app:multi_direction}

\subsection{Component Indexing and Embeddings}
In our framework, the component index $d$ maps to tensor components as follows: $0 \rightarrow D_{xx}$, $1 \rightarrow D_{yy}$, $2 \rightarrow D_{zz}$, $3 \rightarrow D_{xy}$, $4 \rightarrow D_{xz}$, $5 \rightarrow D_{yz}$. The component embedding $\mathbf{e}_d$ is produced by a trainable Fourier embedding module: a learnable vector of size $6 \times 3C$ is retrieved per component index and combined with pre-computed 3D sinusoidal positional encodings via a dot product to yield a spatially aware embedding volume, where $C$ is the latent channel dimension.

\subsection{Shared Anatomical Conditioning}
The shared conditioner $\mathcal{C}$ processes the four input DWI volumes ($b{=}0, b_{1000,x}, b_{1000,y}, b_{1000,z}$) through separate feature extraction pathways, producing per-volume multi-scale feature maps $\{\phi_c^{(l)}\}_l$. These features are used differently in the two phases:
\begin{itemize}
    \item \textbf{Phase~I (Anatomy-Conditioned Autoencoder):} The per-volume features are concatenated and uniformly fused at each scale via a learned multi-stage convolutional network to produce a component-independent conditioning representation $\Phi_{\mathbf{X}}$, which is injected into all component-specific decoders via Feature-wise Linear Modulation (FiLM)~\cite{perez2018film}.
    \item \textbf{Phase~II (Latent Diffusion):} The same per-volume features are concatenated, downsampled to per-volume representations $\bar{\phi}_c$, and fed into the MoE DWI conditioner. The MoE conditioner processes each $\bar{\phi}_c$ through a dedicated expert encoder $f_c$ and applies component-adaptive gating to produce $\Phi^d_{\mathbf{X}}$, which is injected into the denoising U-Net via FiLM.
\end{itemize}

\textbf{Routing specialization.} On the submitted model, the learned gating is only mildly component-dependent: diagonal components place the most weight on the $b_{1000,y}$ expert (average weight 0.40 \vs\ 0.25 for a uniform average, routing entropy 1.23 nats), whereas off-diagonal components weakly favor the $b_{1000,z}$ expert (average weight 0.34, entropy 1.37 nats), against a maximal entropy of 1.39 nats for four experts. The gating thus remains close to uniform but routes diagonal components more sharply than off-diagonal ones, consistent with mild component-adaptive conditioning rather than a strictly uniform average.

\section{Comparison with Coupled Approaches}
\label{app:coupled_comparison}

\Cref{tab:coupled_comparison} highlights the key architectural differences between a standard coupled latent field approach (\eg, stacking all channels) and our component-wise framework.

\begin{table}[H]
\caption{\textbf{Comparison of Coupled \vs\ Component-Wise Architectures.}}
\label{tab:coupled_comparison}
\centering
\resizebox{\columnwidth}{!}{%
\begin{tabular}{lcc}
\toprule
Feature & Coupled Field & Component-Wise (Ours) \\
\midrule
Latent Representation & Joint ($[6{+}4] \times H \times W \times S$) & Independent ($6 \times 1 \times H \times W \times S$) \\
Conditioning & Concatenation & Shared DWI Conditioner + Fusion \\
Cross-Component Modeling & Implicit (Conv layers) & Explicit (Attention) \\
Parameter Efficiency & Shared weights & Shared backbone + Direction Embeddings \\
\bottomrule
\end{tabular}%
}
\end{table}

\section{Limitations}
\label{app:limitations}

All experiments use the HCP Young Adult dataset with uniform, high-quality acquisitions. Generalization to older populations, individuals with neurological conditions, or data from multiple sites remains untested~\cite{madden2012diffusion}. Our experiments use single-shell $b{=}1000$~s/mm$^2$ data at 2\,mm isotropic resolution with a fixed four-volume input, a configuration that differs from multi-shell or high-angular-resolution regimes used for more complex diffusion models~\cite{jeurissen2014multi, alexander2019imaging, tournier2007robust}. TensorLDM's inference cost has two reducible factors. On the A6000 used for \cref{tab:runtime}, the 250-step DDIM sampler takes 123\,s/vol, about five times a vanilla LDM. The bulk is iterative sampling. On a newer RTX 6000 Ada the same 250-step sampler runs in 32.7\,s/vol, and reducing to 25 steps further lowers it to 9.4\,s/vol (\cref{app:runtime}). Thus hardware and step count are independent levers, and a fast sampler mitigates the latency concern for clinical workflows. Because TensorLDM is a stochastic generative model, its output variability across sampling runs has not been characterized. Deterministic or low-variance alternatives may be preferable in safety-critical settings. Finally, the reference tensors are computed using a single fitting algorithm. Evaluating robustness across alternative fitting methods is an important direction for future work~\cite{jones2010twenty}.

\end{document}